\documentclass{article}

\usepackage{arxiv}

\usepackage[utf8]{inputenc} 
\usepackage[T1]{fontenc}    
\usepackage{hyperref}       
\usepackage{url}            
\usepackage{booktabs}       
\usepackage{amsfonts}       
\usepackage{nicefrac}       
\usepackage{microtype}      
\usepackage{lipsum}		
\usepackage{graphicx}
\usepackage[numbers]{natbib}
\usepackage{doi}
\usepackage{comment}
\usepackage{setspace}
\usepackage{graphicx} 
\usepackage{comment}
\usepackage{gensymb}
\usepackage{subcaption}
\usepackage{float}
\usepackage{multirow}
\usepackage{array}
\usepackage{tikz}

\title{Ankle Exoskeletons in Walking and Load-Carrying Tasks: Insights into Biomechanics and Human-Robot Interaction}

\author{ \href{https://orcid.org/0009-0007-7892-582X}{\includegraphics[scale=0.06]{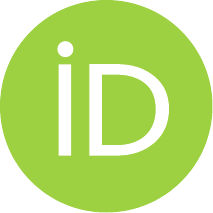}\hspace{1mm}Joana F. Almeida}\thanks{Corresponding author.} \\
	Center for MicroElectroMechanical Systems (CMEMS)\\
	University of Minho, Portugal, Guimarães\\
	\texttt{pg50435@alunos.uminho.pt} \\
        \And
	João C. André \\
	Center for MicroElectroMechanical Systems (CMEMS)\\
	University of Minho, Portugal, Guimarães\\
	\And
	\href{https://orcid.org/0000-0003-0023-7203}{\includegraphics[scale=0.06]{orcid.pdf}\hspace{1mm}Cristina P. Santos} \\
	Center for MicroElectroMechanical Systems (CMEMS)\\
        LABBELS - Associate Laboratory\\
	University of Minho, Portugal, Guimarães\\
	\texttt{cristina@dei.uminho.pt} \\
}

\date{}

\renewcommand{\shorttitle}{\textit{arXiv} Template}

\hypersetup{
pdftitle={A template for the arxiv style},
pdfsubject={q-bio.NC, q-bio.QM},
pdfauthor={David S.~Hippocampus, Elias D.~Striatum},
pdfkeywords={First keyword, Second keyword, More},
}

\begin{document}
\maketitle

\begin{abstract}
Background: Lower limb exoskeletons are in the focus of the scientific community due to their potential to enhance human quality of life across diverse scenarios. However, their widespread adoption remains limited by the lack of comprehensive frameworks to understand their biomechanical and human-robot interaction (HRI) impacts, which are essential for developing adaptive and personalized control strategies. To address this, understanding the exoskeleton's effects on kinematic, kinetic, and electromyographic signals, as well as HRI dynamics, is paramount to achieve improved usability of wearable robots.
Objectives: This study aims to provide a systematic methodology to evaluate the impact of an ankle exoskeleton on human movement during walking and load-carrying (10 kg front pack) tasks, focusing on joint kinematics, muscle activity, and HRI torque signals. The methodology is designed to account for individual and device-specific factors, ensuring adaptability across users and exoskeletons.
Materials and Methods: The study employed an inertial data acquisition system (Xsens MVN), electromyography (Delsys), and a unilateral ankle exoskeleton. Three complementary experiments were performed. The first examined basic dorsiflexion and plantarflexion movements. The second analysed the gait of two subjects without and with the device under passive and active assistance modes. The third investigated load-carrying tasks under the same assistance modes.
Results and Conclusions: The first experiment confirmed that the HRI sensor captured both voluntary and involuntary torques, providing directional torque insights. The second experiment showed that the device slightly restricted ankle range of motion (RoM) but supported normal gait patterns across all assistance modes. The exoskeleton reduced muscle activity, particularly in active mode. HRI torque varied according to gait phases and highlighted reduced synchronisation, suggesting a need for improved support.
The third experiment revealed that load-carrying increased GM and TA muscle activity, but the device partially mitigated user effort by reducing muscle activity compared to unassisted walking. HRI increased during load-carrying, providing insights into user-device dynamics. These results demonstrate the importance of tailoring exoskeleton evaluation methods to specific devices and users, while offering a framework for future studies on exoskeleton biomechanics and HRI.
\end{abstract}


\section{Introduction}

Lower limb exoskeletons and orthoses have been a focal point of the scientific community in recent years, aiming to improve quality of life across medical, military, and industrial applications \citep{cao2021lower, kalita2021development}.
These devices are evolving considering the increasing demand for support, protection, and the enhancement of human performance. Their demonstrated potential to provide effective assistance suggests they will play a significant role in improving daily life for individuals.

A major challenge for wearable robotics is the development of control strategies that provide assistance efficiently and appropriately for each individual \citep{pons2008wearable, sun2022sensing}. 
The human motor system is complex, as movement exhibits unique characteristics for each individual that can even change over time, highlighting the inherent unpredictability of human responses.
For that reason, control strategies of any device that is closely coupled to the human body needs to accommodate the singularities of the dynamic human motor system for preserving the human natural movement patterns and improving usability by ensuring safe and comfortable interactions between the user and the device \citep{zhang2015experimental}.
In order to do so, the human's needs and intentions must be integrated in the closed loop control system, enabling dynamic and continuous adaptations. Designing such control methods requires a deep understanding of the mechanisms of human movement, the device and the interactions between the device and the user, as the power transfer from the robot to the human is inherently shaped by both parties of the system \citep{pons2008wearable}.

The literature contains a variety of studies that delve into human biomechanics across a wide range of tasks, from walking \citep{padulo2023gait} and running \citep{reznick2021lower} to other movements \citep{grimmer2020human}.
In addition, numerous studies have focused on the application of wearable robotics to assess their impact on human movement \citep{bryan2021optimized, malcolm2018bi, cao2021lower, lee2018autonomous}.
These investigations have enhanced the understanding of the kinematics, kinetics, and muscle activation patterns associated with human locomotion. However, devising strategies for a specific device necessitates tailored research that takes into account its unique characteristics and those of its target users.
Previous studies have demonstrated the ability of ankle exoskeletons in assisting walking tasks, focusing on reducing human metabolic cost \citep{bougrinat2019design, galle2017reducing, jackson2019heuristic, malcolm2017varying, mooney2016biomechanical, shafer2021neuromechanics, song2021optimizing, zhang2015experimental} as well as reducing user effort during load-carrying \citep{mooney2014autonomous}.

Based on this body of research, the present biomechanical study employs a different ankle exoskeleton to deliver assistance and aims to investigate its impact on the human locomotion during normal walking and load-carrying tasks. This is achieved through an in-depth analysis of kinematic and electromyographic data.
Conducting this biomechanical study was deemed imperative to gain a deeper understanding of human movement and to assess the effectiveness of the exoskeleton's assistance. 
Furthermore, the findings aim to provide critical insights to guide the development of advanced control strategies for this specific device. Finally, this study indicates the steps that drive the development of effective control strategies for any  wearable robotic device in general.

In addition to determining the impact of the device on kinematics and muscle activity, this study is simultaneously interested in measuring the human-robot interactions (HRI), since the user's movement intentions can be determined from the interaction forces exerted on the device. HRI can be measured directly using force or torque sensors or indirectly through estimators that rely on alternative data sources.
The latter approach has been the most commonly employed in current developments and given the limited research on physical HRI, this study focuses on analysing HRI torque data obtained from the torque sensor embedded in the exoskeleton.

Unlike most studies in the literature on load transport, this research addresses specifically the effects of front packs on the human body. While it is well established that load placement presents distinct consequences on gait, studies examining front pack transport remain limited \citep{yali2012biomechanics}, as the majority concentrate on backpack loading \citep{genitrini2022impact, wang2023effect, kunzler2023effect}. Further exploration into the impacts of carrying loads on the front of the body is necessary, given its prevalence in daily activities, particularly among workers in occupational and industrial environments.

This article presents the findings from an investigation into human biomechanics and Human-Robot Interaction (HRI) using an ankle exoskeleton to assist with two motor tasks: normal walking and load-carrying. 
It is divided into three analyses. 
The first experiment is designed to comprehend the measurements of the HRI torque sensor in basic ankle movements. As the joint moves in the sagittal plane, various forces are applied perpendicular to the motor's axis of rotation, generating torques. The goal is to evaluate how voluntary and involuntary forces exerted by the user on the device are detected by the torque sensor.
The second analysis focuses on evaluating the impact of ankle exoskeleton assistance on human gait. Specifically, it aims to determine whether significant changes occur in ankle kinematics and muscle activity, and also clarify the evolution of HRI torque under dynamic conditions.
Finally, the third experiment intends to explore the biomechanics of the human during loaded walking with a front load. It evaluates changes in movement patterns, muscular requirements and the HRI torque, exploring whether this demanding task alters the user's interaction with the device.


\section{Materials and Methods}

\subsection{Ankle Exoskeleton}

\begin{figure}[htb]
    \centering
    \includegraphics[width=0.8\linewidth]{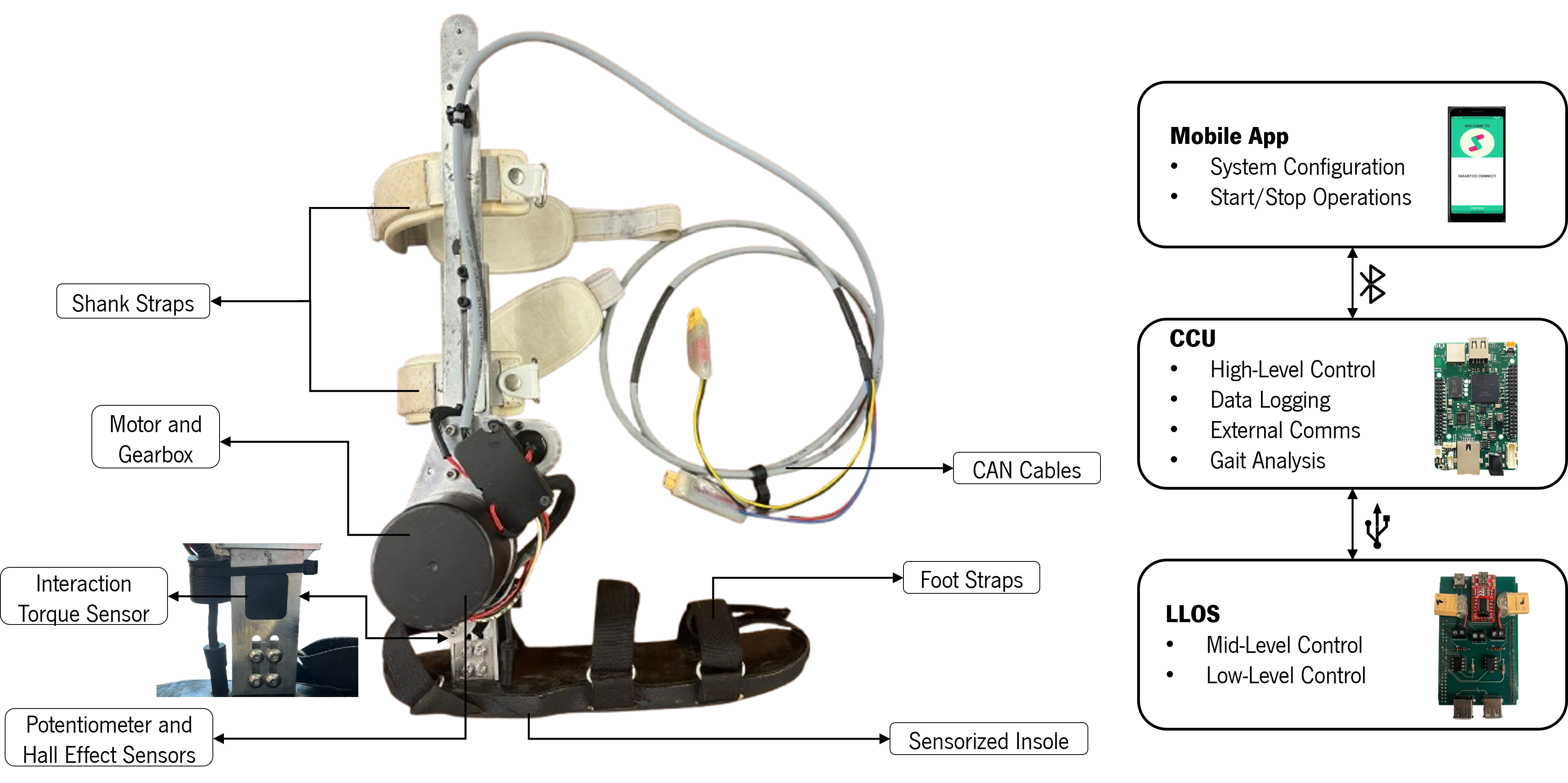}
    \caption{SmartOs architecture and exoskeleton's mechanical structure.}
    \label{fig:SmartOsArch}
\end{figure}

Assistance was provided by the SmartOs system developed at BirdLab, presented in Figure \ref{fig:SmartOsArch}. The system employs a non-centralized architecture with a hierarchical organization, comprising an exoskeleton (Exo-H2 module from Technaid S.L., Spain) with one active DoF for the right ankle joint, a central control unit (CCU), a development board (LLOS), an Android application and a power supply. 

The device's actuator is an electrical motor (EC60-100W Flat Brushless DC Motor, Maxon, Germany) coupled to a gearbox (CSD20-160-2A Strain Wave gear with ratio 160:1, Harmonic Drive, Japan), that can provide average torques of up to 35 Nm and peak torques of 180 Nm. The system is powered by a lithium iron phosphate battery (LiFePO4) of 22.4 V with an estimated autonomy of 8 hours. The mechanical structure was designed to be worn by users with heights ranging from 150 to 190 cm and weights from 45 to 100 kg. The weight of the system is approximately 3 kg. Additionally, the device can assist a range of gait speeds limited from 0.5 to 1.6 km/h. 

SmartOs was developed with a three layered hierarchical control organization: high, middle and low level, as proposed in \citep{tucker2015control}. 
The CCU is a UDOO x86 computer with a quad-core central processing unit (64 bits, 2.24 GHz and 4 GB of RAM) operating a Linux system and running the higher-level controls at 100 Hz.
The LLOS is an STM32F4-Discovery board (STMicroelectronics, Switzerland) coupled with USB for external communications and runs the mid and low level control layers at 100Hz and 1kHz, respectively. 
SmartOs framework comprises a Mobile App developed in Android Studio for personalized configuration of the control and start/stop commands.

The CCU communicates with the Mobile App via Bluetooth and interfaces with the LLOS via USB by UART communication protocol. The high-level program in the CCU was developed in C++ using the software QtCreator, whilst the LLOS STM32F4 board was programmed in C using freeRTOS operating system in Keil uVision 5.0 IDE. The LLOS board establishes direct communication with the driver of the ankle exoskeleton through CAN protocol.

SmartOs currently endows two control modes: passive and active position.
Passive control consists of a torque-based PID control that minimizes the mechanical impedance of the device by following the user's motion intentions, detected through the forces applied by the user in the device through the straps.
Active control consists of a position-based PID controller designed to follow a predefined walking gait cycle trajectory, based on the error between the reference position and the angle of the exoskeleton.

\subsection{Measured Outcomes}

\subsubsection{Exoskeleton Mechanics and Human-Robot Interaction}
The ankle exoskeleton's on-board sensors were utilised to measure the device's kinematics and kinetics. The exoskeleton's angular trajectory was recorded using a built-in precision potentiometer. A Hall effect sensor within the device was used to calculate the torque exerted by the DC motor. Additionally, the exoskeleton's insole is equipped with two force-sensing resistors (FSRs) located at the heel and toe areas, used to measure ground contact forces.
A strain gauge sensor, positioned on the internal bar segment of the exoskeleton below the motor, as in Figure \ref{fig:SmartOsArch}, and configured in a full Wheatstone bridge, measured the human-robot interaction torques.

\subsubsection{Lower Limb Movement}

For assessment of human lower limb join kinematics/kinetics, the MTw Awinda motion tracking system was used. 
Each participant was fitted with Xsens Awinda IMUs configured for lower limb tracking. Seven sensors were strategically placed on the body: one on each foot, each lower leg, and each thigh, as well as one on the pelvis (Appendix \ref{app1} for visual placement). Calibration was performed using the Xsens MVN software (version 2021.2), following the software's standard instructions to ensure precise movement tracking.
The software processes the raw sensor data to extract the angles of all lower-limb joints and the contact data of both feet with the ground.

\subsubsection{Ankle Muscle Activity}

Two surface electromyography (sEMG) sensors from the Trigno Wireless EMG System (Delsys Inc., Natick, MA, USA) were used to monitor the muscles responsible for ankle movements.
The Trigno system was integrated into the SmartOs framework, enabling synchronized acquisition of EMG envelope signals during locomotion via the Trigno SDK. This integration facilitated the calculation of Maximum Voluntary Contraction (MVC) values and the collection of real-time muscle activation data.

As the study focused on sagittal plane movements—specifically dorsiflexion (DF) and plantarflexion (PF)—the Tibialis Anterior (TA) and Gastrocnemius Medialis (GM) muscles were selected for monitoring due to their high activity during these movements and their accessible peripheral locations. The placement of the two sEMG sensors can be consulted in Appendix \ref{app1} and followed the SENIAM recommendations for sensor locations in lower leg or foot muscles \citep{hermens2000development}.

\subsection{Data Collection}

The collection of data was divided into three experiments, each addressing specific research questions.
The first protocol comprises the execution of two basic movements of the ankle joint.
The second protocol collects data from individuals walking on a treadmill and the third was outlined for load-carrying tasks.

All participants involved in data collection were familiar with SmartOs, having previously used it in both control modes over multiple sessions.
A treadmill was used to ensure a consistent walking speed of 1.3km/h and the participants were instructed to hold the treadmill handrails solely for balance, avoiding using the upper limbs for weight support.

\begin{figure}[htb]
    \centering
    \includegraphics[width=0.8\linewidth]{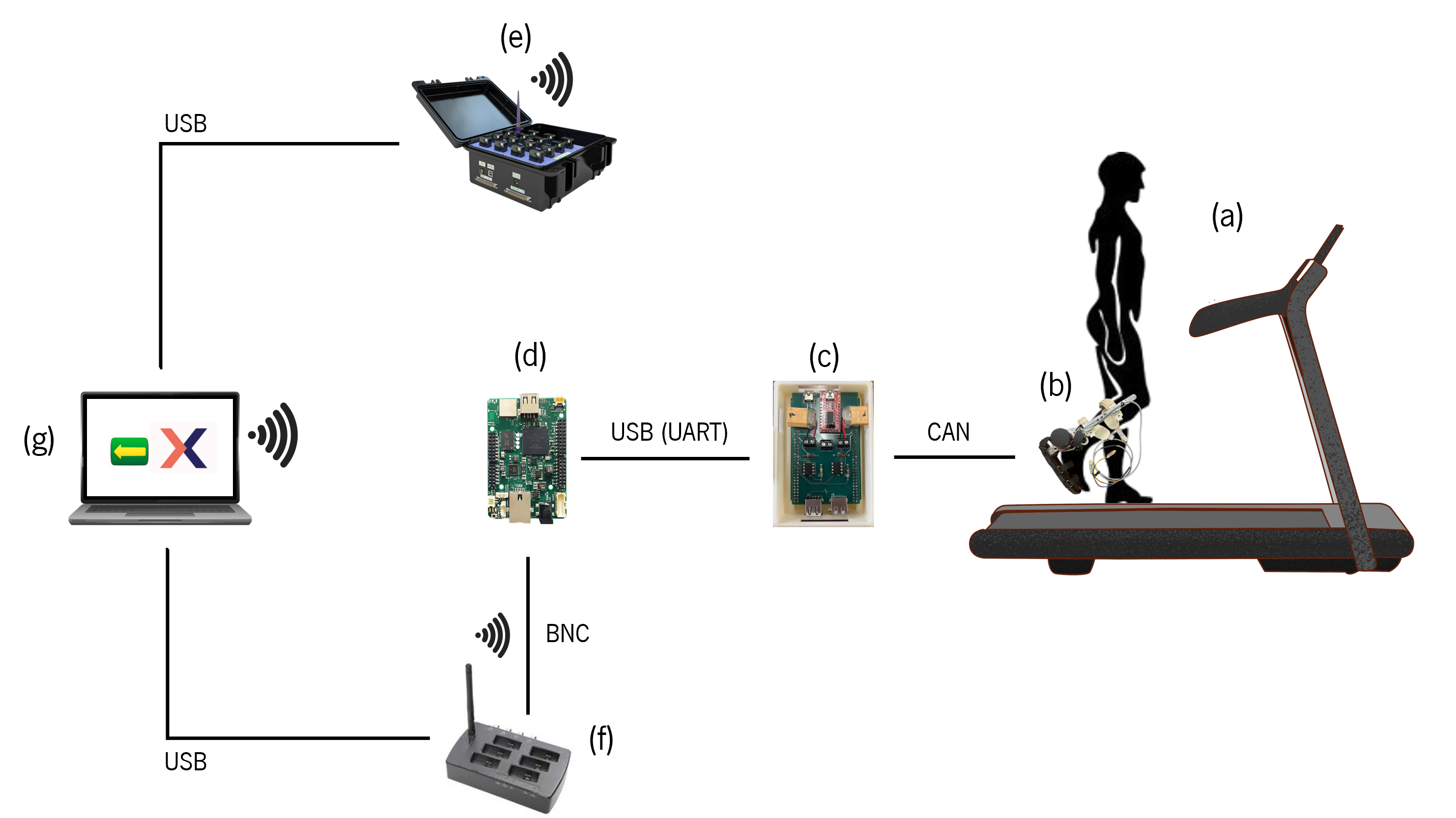}
    \caption{Experimental setup for walking data collection. (a) Treadmill. (b) Ankle-H2 module. (c) LLOS board. (d) CCU board. (e) Delsys Trigno Workstation with wireless communication with the Delsys sensors. (f)  Xsens Awinda station with wireless communication with the Xsens IMUs. (g) Laptop running Trigno Control Utility and Xsens MVN 2021.2 with wireless communication with CCU.}
    \label{fig:experimentalsetup}
\end{figure}

Data collection for the first protocol relied exclusively on the exoskeleton's integrated sensors, whereas the other two protocols utilized the MVN and Trigno EMG systems synchronized with SmartOs.
EMG sensors were paired with the base station using the Trigno Control Unit software and data was transmitted via Wi-Fi to the SmartOs CCU and stored separately for each trial.
The MVN system was synchronised with SmartOs through a trigger box. At the start of each trial, SmartOs sent a trigger to MVN system to start inertial data recording, which was subsequently stored on a local computer.
The experimental setup, including all systems and their integration, is illustrated in Figure \ref{fig:experimentalsetup}.

All data recording was conducted at the BirdLab's facility, at University of Minho, under the ethical procedures of the Ethics Committee in Life and Health Sciences (CEICVS 006/2020), following the Helsinki Declaration and the Oviedo Convention.

\subsection{Data Processing} 

All data was sampled at 100 Hz. After acquisition, it was imported into MATLAB 2022b (Mathworks, Natick, MA, USA) for processing and analysis. The results reported in this study are of the right leg only.

Data processing involved several key steps.
First, the collected data was segmented into individual strides. For trials without the exoskeleton, heel strike (HS) and toe-off (TO) events were detected using the Xsens MVN foot contact data. In trials involving the exoskeleton, an algorithm was developed to identify gait events based on the force-sensing resistor (FSR) data. To validate this algorithm, the FSR-detected events were cross-referenced with Xsens foot contact data and the angular velocity of the foot, ensuring accurate segmentation.
Once segmented, all strides were normalised to 500 samples using linear interpolation and expressed as percentages of the gait cycle. Outlier strides were excluded using a threshold of three standard deviations.

For each trial, kinematic, kinetic, and EMG data were averaged per participant. Kinematic metrics, including maximum and minimum angles, stance phase duration, and initial contact angles, were calculated for each stride and subsequently averaged.
The EMG data, saved as linear envelopes after filtering and rectification, were further normalised using MVC. For each muscle, maximum activation values and cumulative activity over specific gait phases were determined for comparison.
Processing of the HRI torque data included normalisation by the participant’s body weight, and calculation of the mean HRI torque for each condition. Additionally, cumulative HRI torque over the gait cycle was computed, enabling comparison across trials.

No statistical analysis was performed in this study due to the limited quantity of subjects. Instead, the analysis focused on a purely descriptive and qualitative interpretation of the collected data. This approach enabled an in-depth examination of the individual’s responses to varying conditions, facilitating the identification of trends and patterns without statistical inference. By emphasizing qualitative aspects, the study aims to explore the ankle joint kinematics, muscle activity, and HRI torque in a detailed and context-specific manner.

\section{Results}

\subsection{Basic Ankle Movements}

The first experiment aims to investigate how different foot movements are reflected in the interaction torque measurements.
Given that HRI is influenced by both the user and the device, the analysis included data collected under two control conditions: passive control, where movement is solely induced by the user, and active control, where movement arises either from the device alone or from a combination of the user's and the device's efforts.

\subsubsection{Passive Mode}
\begin{figure}[htb]
    \centering
    \includegraphics[width=0.95\linewidth]{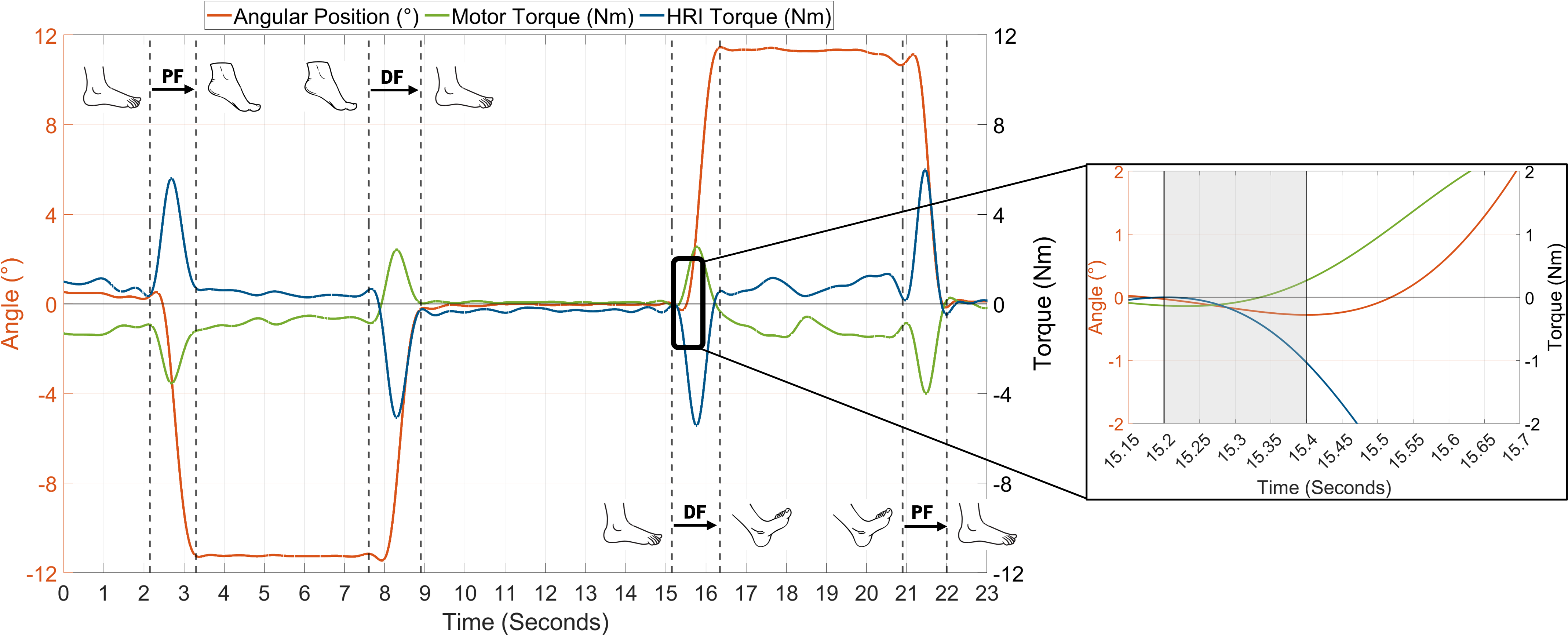}
    \caption{Ankle exoskeleton's angle (orange), motor torque (green) and HRI torque (blue) for DF and PF movements in passive control.}
    \label{fig:P1Passive}
\end{figure}

Figure \ref{fig:P1Passive} demonstrates the results collected with the exoskeleton in passive control mode.
It was found that whenever the person pushed the device in the direction of DF (increasing the angle - orange line), the measured HRI torque (blue line) showed negative HRI peaks. On the other hand, for PF (reducing the angle), the measured HRI torque showed positive values.
Additionally, a causal relationship was identified between the HRI torque and the angular position. An increase in the magnitude of the HRI torque was observed first, followed by a subsequent change in the device's angular position.

\subsubsection{Position Mode}

In active control, when the exoskeleton performed DF from 0\degree to 12\degree, the data read by the torque sensor was as follows: 1) if the user was passive, the HRI torque detected was positive; 2) if the user was opposing, the HRI torque was also positive but reached higher intensities compared to the previous condition; 3) if the user was helping in same direction, the HRI torque measured was negative.
For DF from -12\degree to 0\degree, the difference from the previous results was observed in the passive condition, where the HRI torque was negative instead of positive.

For PF in active mode, moving from 0\degree to -12\degree, the data showed: (1) if the user was passive, the HRI torque was negative; (2) if the user opposed the exoskeleton’s movement, the torque remained negative but exhibited higher intensity; (3) if the user helped the exoskeleton’s PF movement, then the intentional force generated a positive HRI torque.
Finally, during PF from 12\degree to 0\degree, if the user’s foot remained passive, the HRI torque was positive.
The main outcomes from this analysis are expressed in Table \ref{tab:HRIMTSigns} which contains the sign of the measured HRI torque (HRI) and the motor torque (MT) in the four movements.

\begin{table}[htb]
    \centering
    \footnotesize
    \caption{HRI torque (HRI) and motor torque (MT) signs for different DF and PF movements with the exoskeleton in active position control}
    \label{tab:HRIMTSigns}
    \begin{tabular}{cccccc}
    \hline
    \multicolumn{2}{c}{\multirow{2}{*}{\textbf{Movement}}} & \multicolumn{2}{c}{\textbf{DF}}       & \multicolumn{2}{c}{\textbf{PF}}      \\ \cline{3-6} 
    \multicolumn{2}{c}{}                                   & \textbf{[0 +12]\degree} & \textbf{[-12 0]\degree} & \textbf{[0 -12]\degree} & \textbf{[12 0]\degree} \\ \hline
    \multirow{2}{*}{\textbf{Passive}}  & \textbf{MT}  & +  & +  & -  & -  \\ \cline{2-6} 
                                       & \textbf{HRI} & +   & -   & -   & +   \\ \hline
    \multirow{2}{*}{\textbf{Opposing}} & \textbf{MT}  & + & + & - & - \\ \cline{2-6} 
                                       & \textbf{HRI} & +  & +  & -  & -  \\ \hline
    \multirow{2}{*}{\textbf{Helping}}  & \textbf{MT}  & +   & +   & -   & -   \\ \cline{2-6} 
                                       & \textbf{HRI} & -  & -  & +  & +  \\ \hline
    \end{tabular}
\end{table}

\subsection{Ankle Walking Assistance}

This experiment aims to understand how the ankle exoskeleton assists the user during a complex motor task, i.e. walking, comparing the two control modes (passive and active) to walking without exoskeleton assistance (unassisted). 

For this purpose, the analysis focuses on several key objectives. Firstly, the kinematics of the human ankle joint are assessed to verify whether the user is able to reproduce a normal walking with the device on. Second, an evaluation of the effects of the exoskeleton in muscle activity during the locomotive tasks is performed. Lastly, an analysis of the HRI torque is conducted to investigate the forces applied between the human and the device in dynamic conditions, exploring the insights it provides into the user's comfort and assistance needs.

By comparing the three walking conditions - unassisted, passive and active position control- this experiment seeks to determine whether the exoskeleton can effectively assist the ankle joint during gait.

\subsubsection{Kinematics}
\begin{figure}[htb]
    \centering
    \begin{subfigure}[t]{0.55\textwidth}
        \centering
        \includegraphics[width=\linewidth]{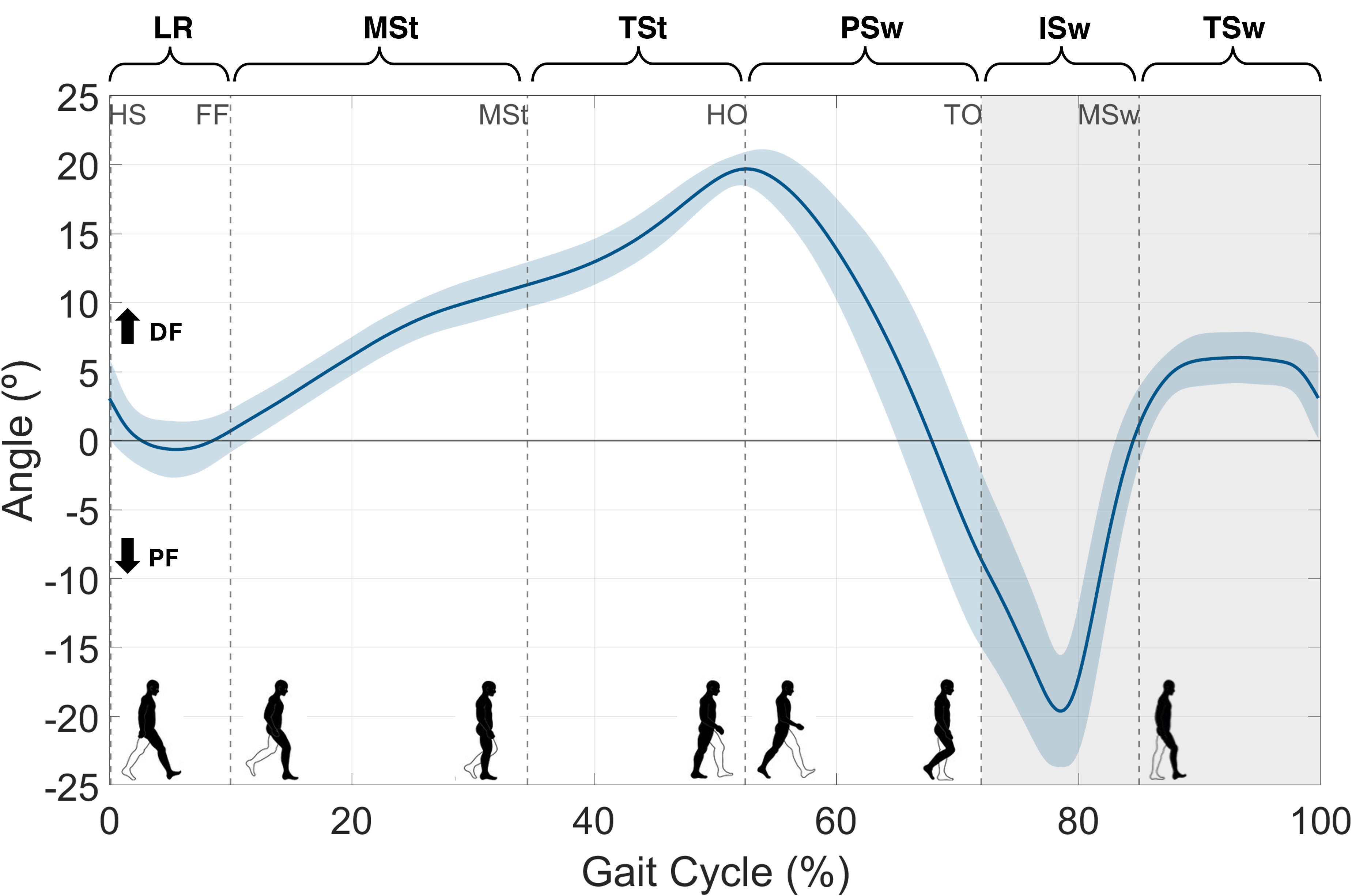}
        \caption{No Assistance.}
        \label{fig:AnkleAngleDuringWalking}
    \end{subfigure}
    \vspace{0.1cm}
    \begin{subfigure}[t]{0.45\textwidth}
        \centering
        \includegraphics[width=\linewidth]{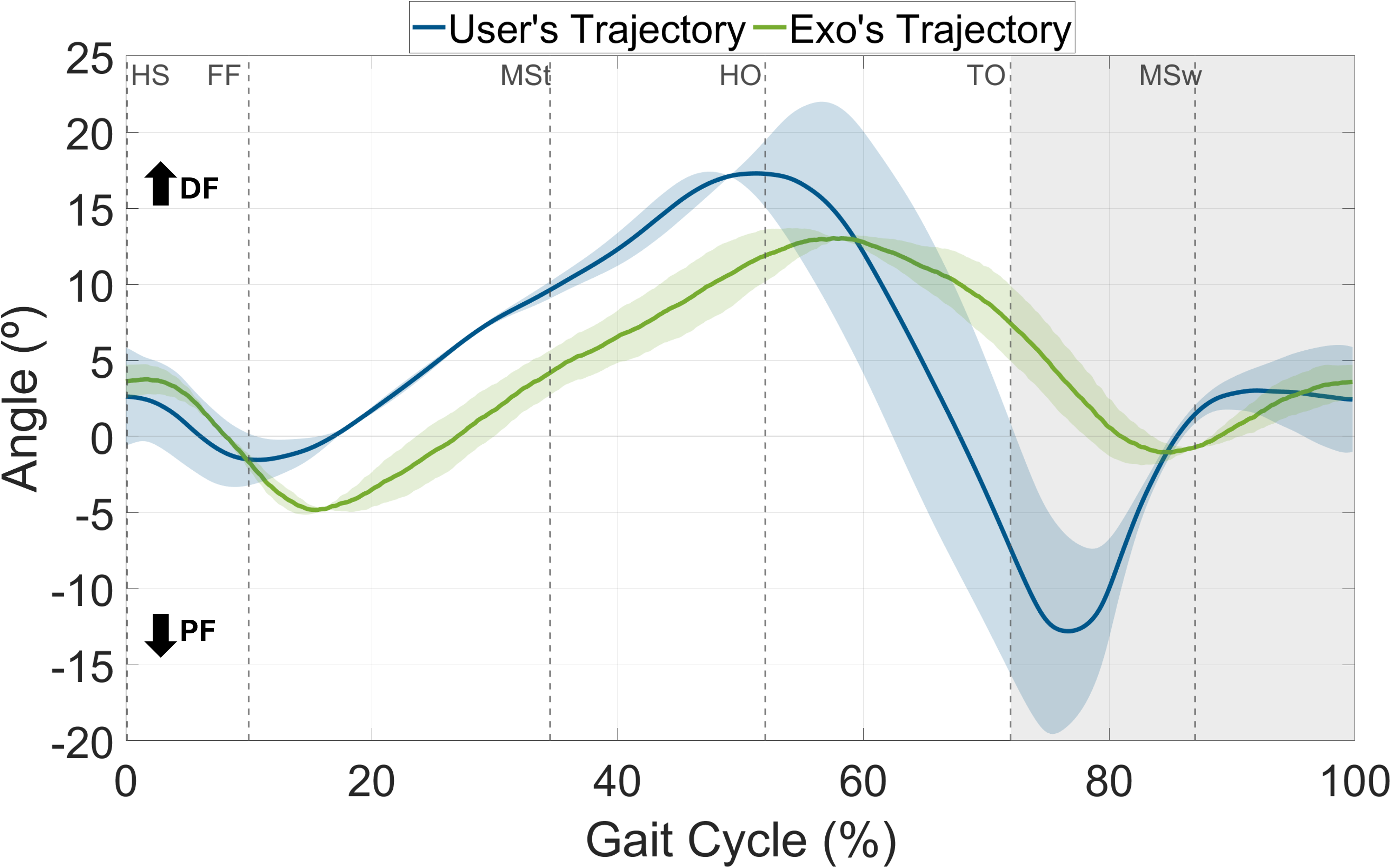}
        \caption{Passive Assistance.}
        \label{fig:AngleFootExoPassive}
    \end{subfigure}
    \hspace{0.1cm}
    \begin{subfigure}[t]{0.45\textwidth}
        \centering
        \includegraphics[width=\linewidth]{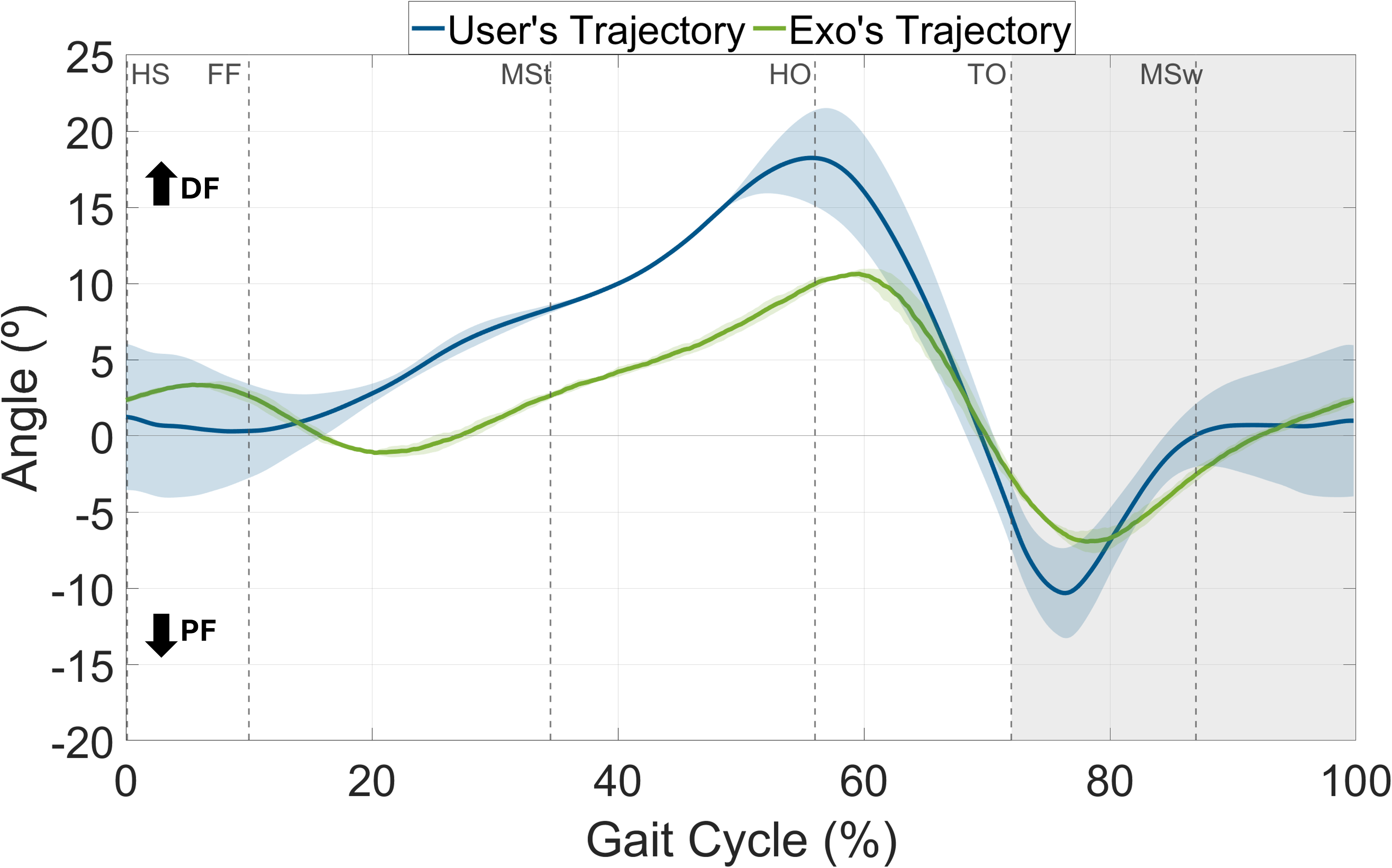}
        \caption{Active Assistance.}
        \label{fig:AngleFootExoPosition}
    \end{subfigure}
    \caption{Angular position of the right ankle joint during one gait cycle, comparing different assistance conditions. The shadowed region represents the swing phase.}
    \label{fig:AngleFootExoCompareModes}
\end{figure}

The analysis of the ankle kinematics of walking at a constant speed of 1.3km/h without any locomotion assistance device is presented in Figure \ref{fig:AnkleAngleDuringWalking}. It contains the six key events of the gait cycle, indicated by dashed vertical lines: Heel Strike (HS), Foot Flat (FF), Middle Stance (MSt), Heel Off (HO), Toe Off (TO), and Middle Swing (MSw). At the top, the gait phases are shown: Loading Response (LR), Middle Stance (MSt), Terminal Stance (TSt), Pre-Swing (PSw), Initial Swing (ISw), and Terminal Swing (TSw).

Figure \ref{fig:AngleFootExoCompareModes} presents the angle of the user's foot (blue solid line) and the ankle exoskeleton's (green solid line) angles when walking in passive (Fig. \ref{fig:AngleFootExoPassive}) and active position control modes (Fig. \ref{fig:AngleFootExoPosition}). Focusing solely on the subject's ankle angle, the observed pattern while walking with the assistive device in any control mode is similar with that noted during unassisted walking. 
However, Figure  \ref{fig:AngleFootExoCompareModes} illustrates that the exoskeleton's Range of Motion (ROM) is consistently lower than that of the user's ankle joint across all control modes.

Regarding the synchronization of movements, in passive assistance, it is observed that the exoskeleton always lags behind the movements of the user. In this control mode, the device fails to respond promply to the movement imposed by the user, especially from HO to MSw.
In active assistance, there is better synchronization throughout the gait cycle overall. However, a mismatch between the exoskeleton and the user occurs during the initial phase of the gait cycle (0–20\%). 

\begin{table}[htb]
    \centering
    \footnotesize
    \caption{Descriptive analysis of kinematics of the human ankle joint in three walking conditions: NA - No Assistance, PA - Passive Assistance, and AA - Active Assistance}
    \label{tab:DescriptiveAnalysisAngPos}
    \begin{tabular}{cccc}
    \hline
                          & \textbf{NA} & \textbf{PA} & \textbf{AA} \\ \hline
    \textbf{ROM (\degree)}          & 41.87 (4.82)    & 34.56 (3.99)     & 30.87 (0.97)      \\
    \textbf{Max DF (\degree)}       & 20.17 (1.05)    & 20.42 (2.62)     & 19.10 (2.86)      \\
    \textbf{Max DF (GC \%)} & 52.66 (1.80)    & 52.06 (4.59)     & 54.89 (2.29)      \\
    \textbf{Max PF (\degree)}       & -21.79 (3.21)   & -14.48 (6.93)    & -10.19 (5.53)     \\
    \textbf{Max PF (GC \%)} & 78.72 (1.37)    & 77.59 (1.85)     & 76.32 (0.22)      \\
    \textbf{Angle IC (\degree)}     & 3.06 (2.67)     & 2.62 (3.57)      & 1.24 (4.80)       \\
    \textbf{Stance (\%)}  & 72.26 (1.63)    & 70.42 (2.28)     & 70.18 (2.24)      \\ \hline
    \end{tabular}
\end{table}

Table \ref{tab:DescriptiveAnalysisAngPos} provides a descriptive analysis of the participant's ankle joint kinematics across the three study conditions.
The participants's ankle joint RoM was reduced by the use of the exoskeleton.
The maximum DF angle showed no significant differences across the three conditions, as indicated by the standard deviations.
On the other hand, PF was restricted by the use of the exoskeleton.
There was a high degree of variability between subjects in executing PF movements, especially in passive control mode. One participant exhibited difficulty performing PF in both passive assistance and active control, achieving maximum angular positions of -7.53\degree and -7.94\degree, respectively.
In contrast, the other participant reached greater PF angles, more limited under active control compared to passive assistance (-20.76\degree in passive and -15.59\degree in active control).
The initial contact angle exhibited variability both within subjects, as it fluctuated across strides, and between subjects.
Regarding stance phase duration, walking without the assistive device showed a slightly longer stance phase (1\% increase) compared to both conditions with the exoskeleton.

\subsubsection{Muscle Activity}

\begin{figure}[htb]
    \centering
    \includegraphics[width=0.7\linewidth]{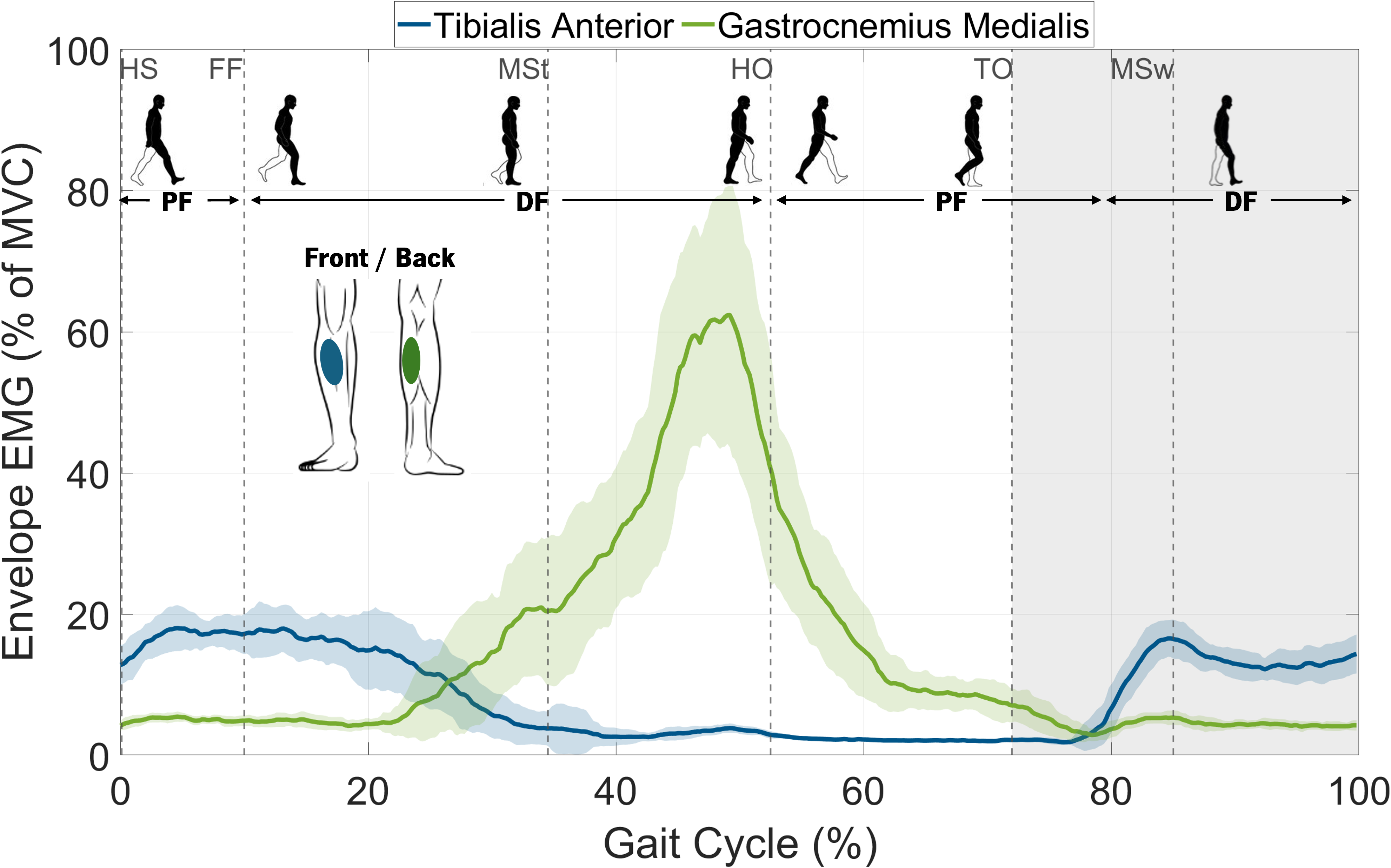}
    \caption{Average gait cycle of muscle activity of TA (blue) and GM (green) of the two participants. The shadow region marks the swing phase.}
    \label{fig:EMGNoExo}
\end{figure}

Figure \ref{fig:EMGNoExo} illustrates the mean muscle activation patterns of the shank muscles throughout the gait cycle during unassisted walking, obtained from both subjects.

It reveals that the muscle activity is complementary. TA is active from HS to MSt and during swing.
In contrast, GM begins its activation around 20\% of the GC and reaches its peak just before HO, around 50\% of the gait cycle.
Comparing the magnitudes of the activation between both muscles during unassisted walking, it is verified that GM typically exceeds 50\% of its MVC during walking, whereas TA generally remains around 20\%. 

\begin{figure}[htb]
    \centering
    \begin{subfigure}{0.49\textwidth}
        \centering
        \includegraphics[width=\linewidth]{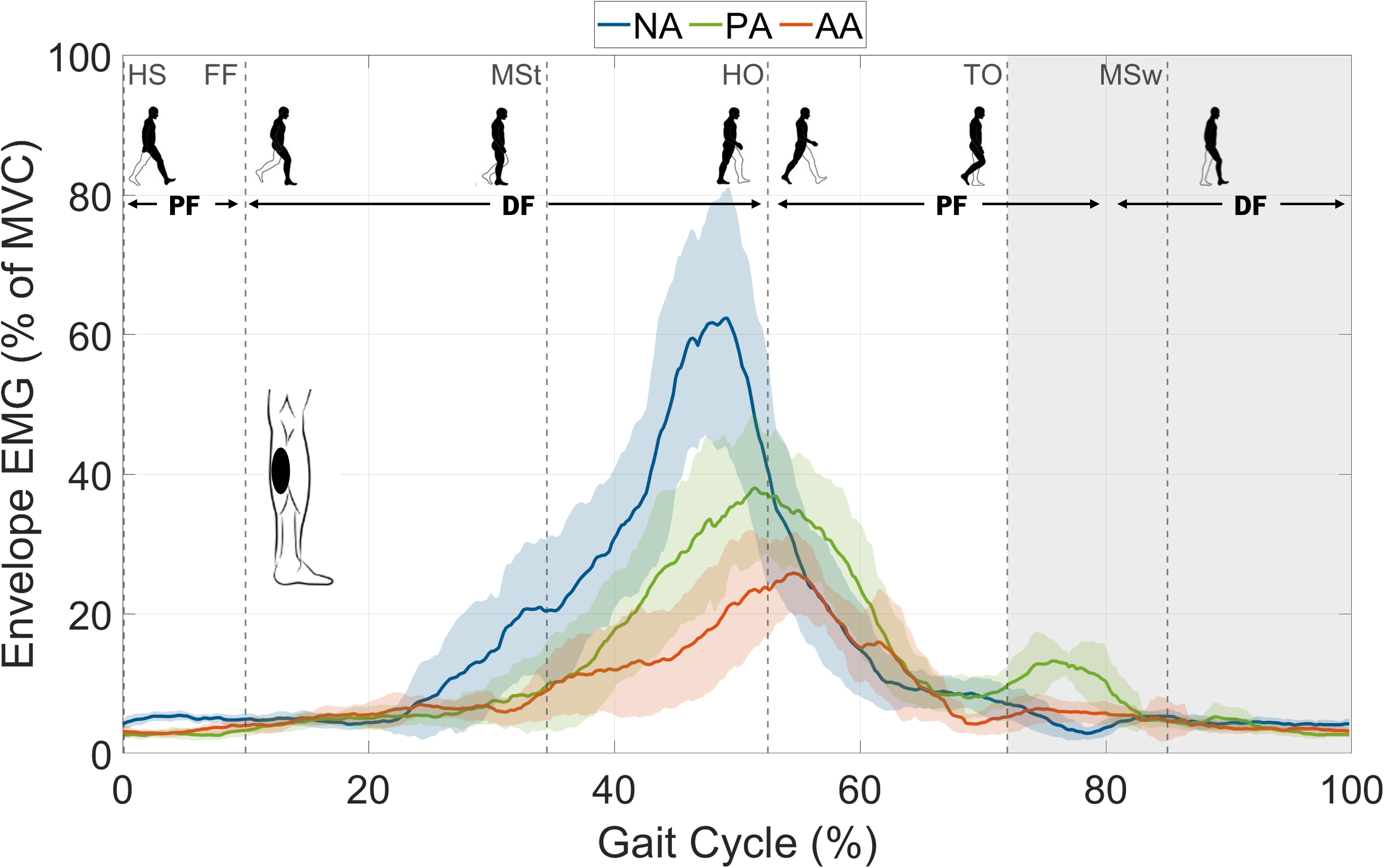}
        \caption{GM.}
        \label{fig:GMCompareAssistance}
    \end{subfigure}
    \hfill
    \begin{subfigure}{0.49\textwidth}
        \centering
        \includegraphics[width=\linewidth]{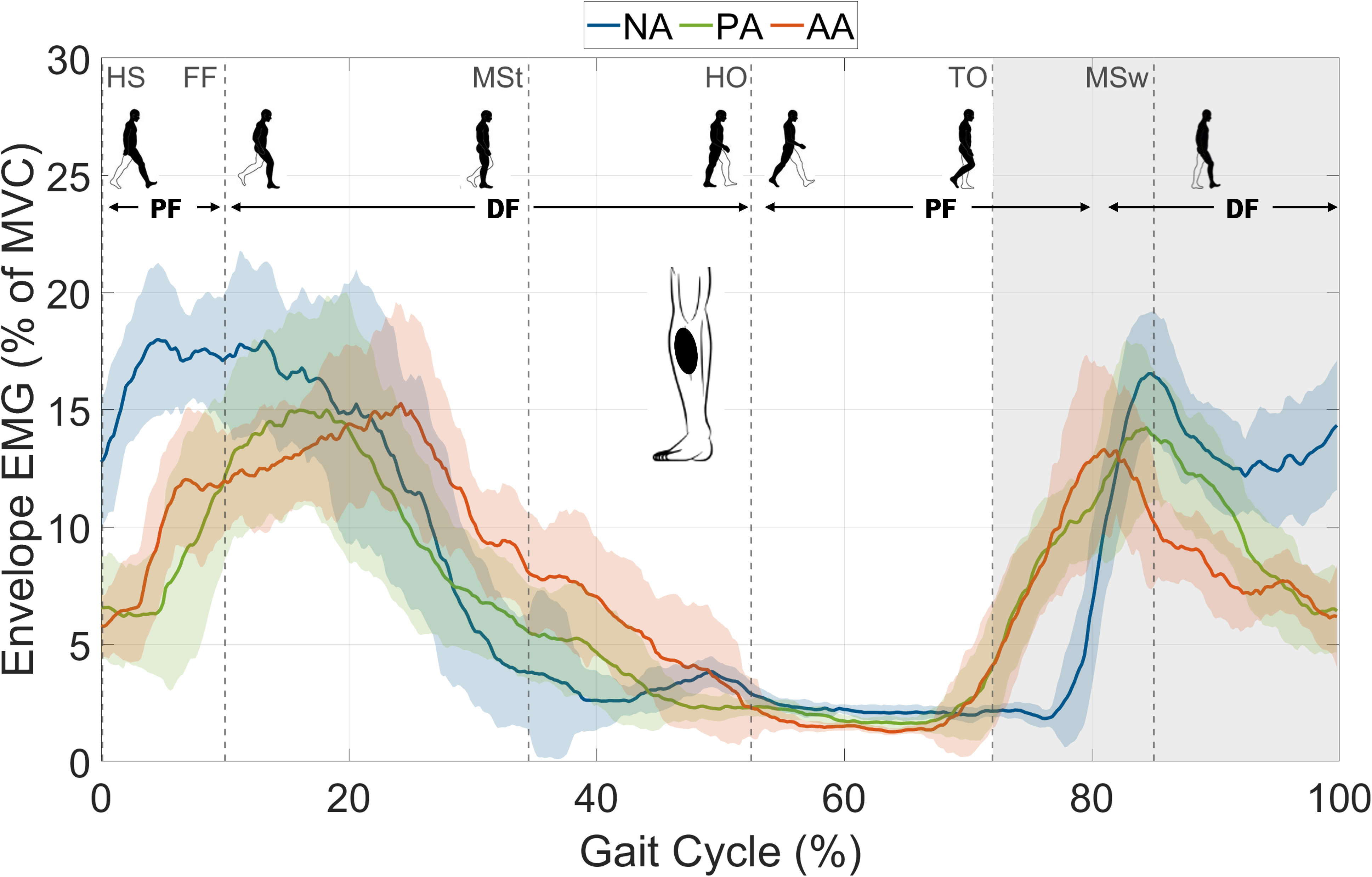}
        \caption{TA.}
        \label{fig:TACompareAssistance}
    \end{subfigure}
    \caption{Envelope of EMG of shank muscles in average gait cycle in three walking conditions: NA - No Assistance, PA - Passive Assistance, and AA - Active Assistance. The shadow region marks the swing phase.}
    \label{fig:EMGCompareAssistance}
\end{figure}

Figure \ref{fig:EMGCompareAssistance} represents the profile of both muscles in the three walking conditions under study.
The data reveals that the activation patterns of both muscles during assistive walking closely mirror those observed during walking without the exoskeleton.

Figure \ref{fig:GMCompareAssistance} shows a peak GM activation around 50\% of the gait cycle across all three conditions, with the highest value occurring during unassisted walking (60\% of MVC), followed by walking with passive assistance (40\% of MVC), and the lowest activation observed during active assistance (25\% of MVC).
In contrast, Figure \ref{fig:TACompareAssistance}, illustrating TA activation across the three conditions, reveals similar muscle activation patterns with maximum values between 15-20\% of MVC.

\begin{figure}[htb]
    \centering
    \begin{subfigure}{0.49\textwidth}
        \centering
        \includegraphics[width=\linewidth]{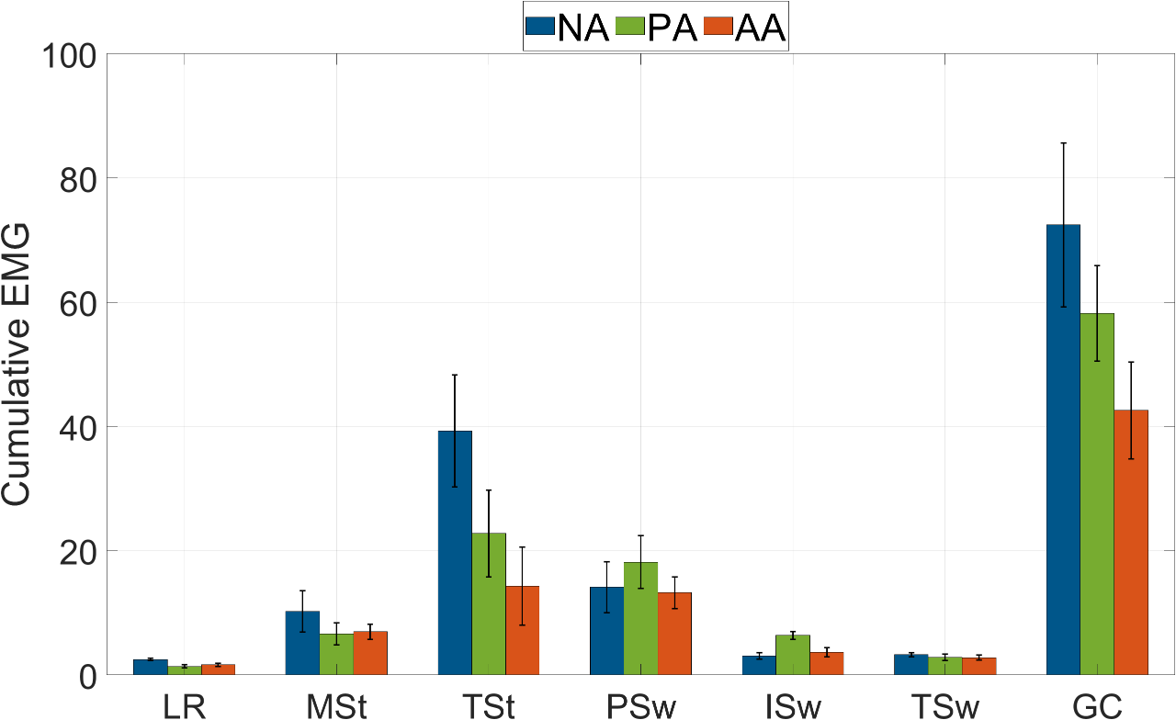}
        \caption{GM.}
        \label{fig:GMCumCompareModes}
    \end{subfigure}
    \hfill
    \begin{subfigure}{0.49\textwidth}
        \centering
        \includegraphics[width=\linewidth]{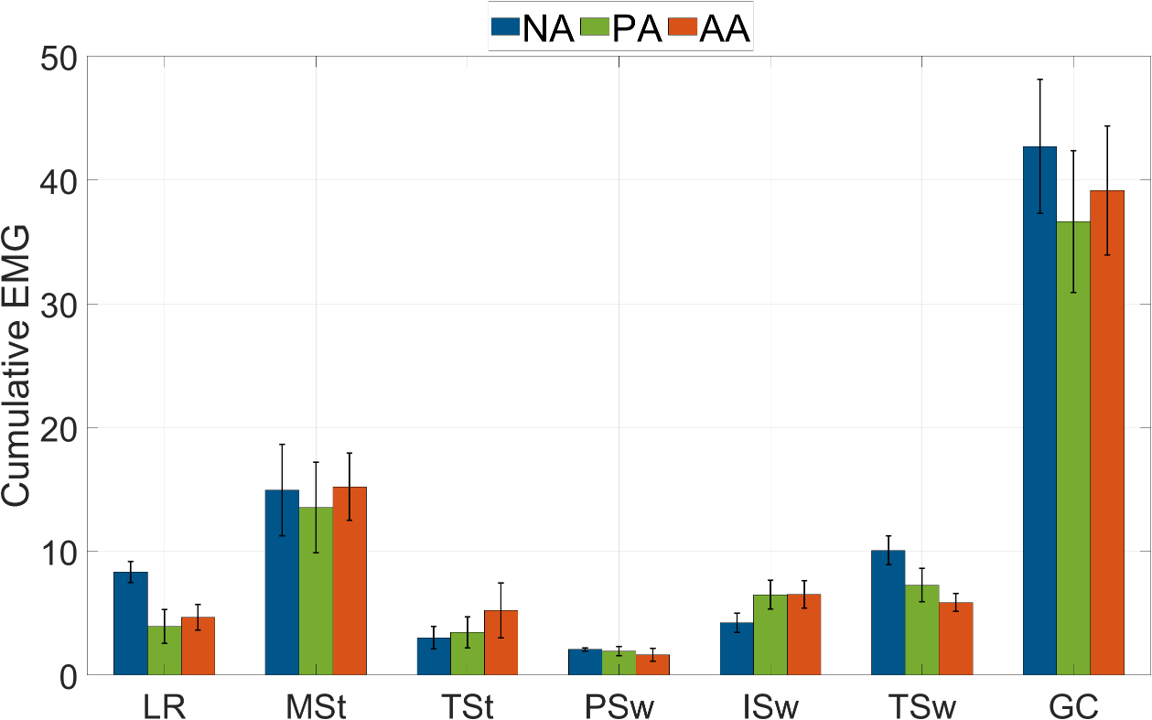}
        \caption{TA.}
        \label{fig:TACumCompareModes}
    \end{subfigure}
    \caption{Comparison of cumulative muscle activity during six subphases of walking and during the complete gait cycle. NA - No Assistance; PA - Passive Assistance; AA - Active Assistance; LR - Loading Response; MSt - Middle Stance; TSt - Terminal Stance; PSw - Pre-Swing; ISw - Initial Swing; TSw - Terminal Swing; GC: Gait Cycle.}
    \label{fig:CumCompareModes}
\end{figure}

Figure \ref{fig:CumCompareModes} presents the cumulative data of the muscle activity for both muscles during the 6 gait subphases and for the complete gait cycle in the three walking modes. 
In Figure \ref{fig:GMCumCompareModes}, the phase of gait that exhibits the most pronounced difference in activation between conditions is the TSt phase, during which GM reaches its peak activation.
The results reveal that the unassisted condition showed the highest cumulative GM activation throughout the gait cycle, compared to the other conditions. It can be verified that the use of the robotic device reduced GM activation, with active control causing the greater reduction.

Regarding TA, Figure \ref{fig:TACumCompareModes} indicates a higher activation in unassisted walking during the first and last subphases of gait (LR and TSw) compared to assistive walking.
During MSt, TA presents higher activity compared to the other gait phases but no significant difference is noted between walking conditions.
The cumulative EMG of TA activity throughout the gait cycle showed that the highest value was recorded during normal walking, followed by active position assistive walking and then passive assistive walking.

\subsubsection{Human-Robot Interaction}

\begin{figure}[htb]
    \centering
    \begin{subfigure}{0.49\textwidth}
        \centering
        \includegraphics[width=\linewidth]{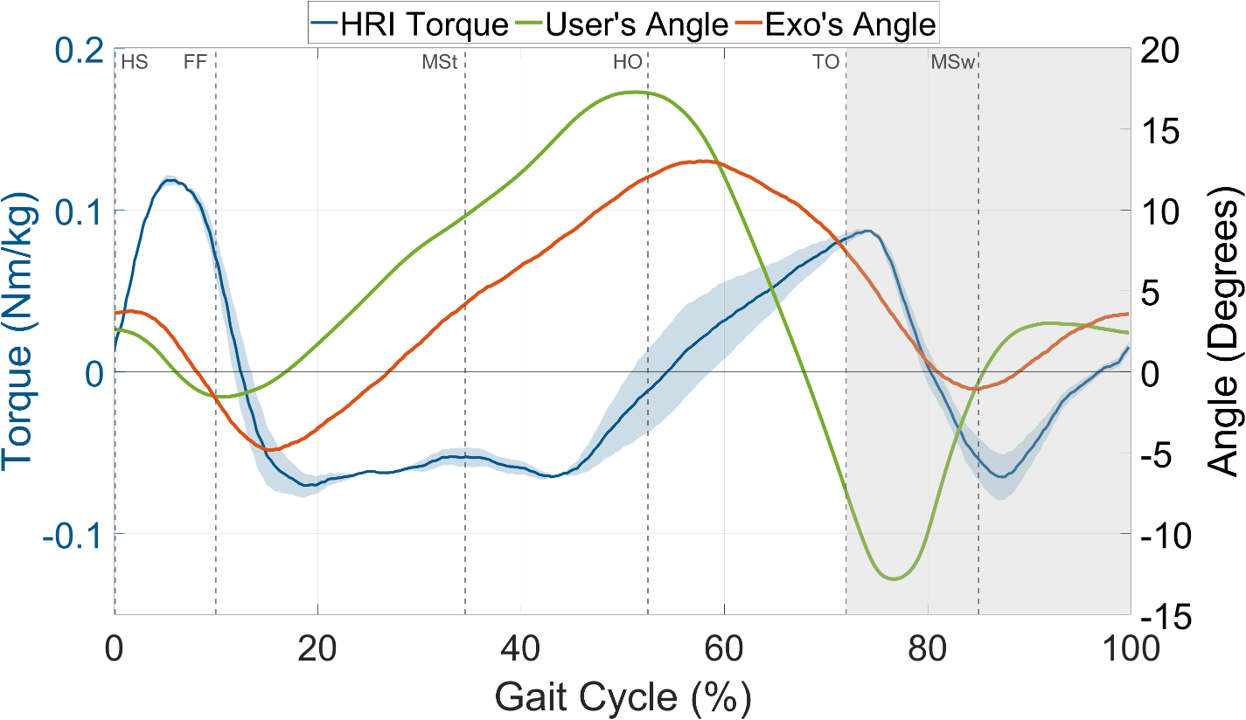}
        \caption{Passive assistance.}
        \label{fig:HRIWalkingPassiveMode}
    \end{subfigure}
    \hfill
    \begin{subfigure}{0.49\textwidth}
        \centering
        \includegraphics[width=\linewidth]{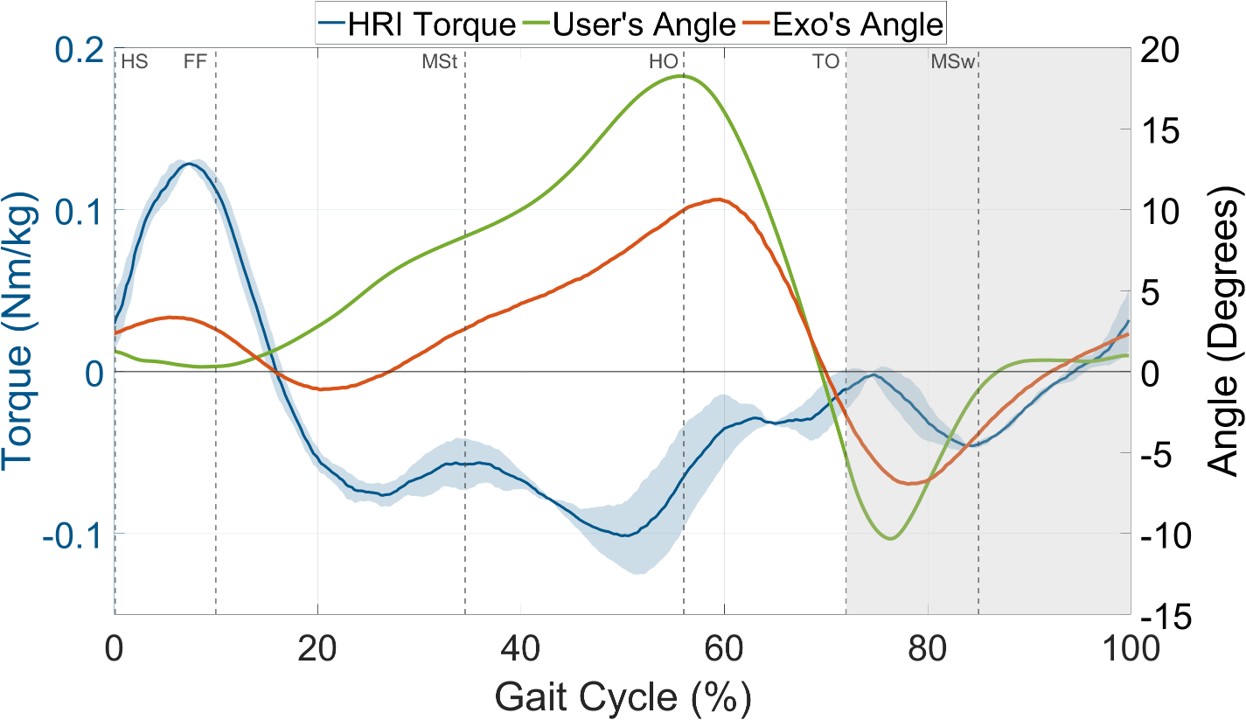}
        \caption{Active assistance.}
        \label{fig:HRIWalkingPositionMode}
    \end{subfigure}
    
    \caption{Average gait cycle of HRI Torque (blue), user's angle (green) and exoskeleton's angle (orange) for two walking assistance conditions.}
    \label{fig:HRICompareModes}
\end{figure}

Figure \ref{fig:HRICompareModes} illustrates the measured HRI torque (blue solid line) throughout the gait cycle when walking with the exoskeleton in passive (Fig. \ref{fig:HRIWalkingPassiveMode}) and active control settings (Fig. \ref{fig:HRIWalkingPositionMode}).
The angular trajectories of the subjects' right foot is the green line and the exoskeleton's is the orange line.

During passive assistance, the HRI torque displays a positive peak at the beginning of the gait cycle, indicating the user forces the device into PF. From FF to HO, the HRI torque is negative, since the user performs DF. After HO, the user progressively pushes the device into PF, reaching a second positive HRI peak at around TO. In swing phase, the HRI torque changes from positive to negative until MSw and approaches zero at the end of the cycle.
During active control, the HRI torque profile similarly features a positive peak during the initial stance phase, followed by a negative interactions, similar to passive control. 
However, from HO to TO, the HRI is closer to zero and negative, indicating that there is no need for the user to actively push the device to achieve PF.
During the swing phase (shaded area), the HRI torque is predominantly negative and follows a similar trend to passive mode.
In addition, Figure \ref{fig:HRIWalkingPositionMode} indicates that as the trajectory of the subject's foot and the device become more aligned, the HRI torque approaches 0 Nm/kg.
Conversely, when the difference between the trajectories increases or when they move in opposite directions, the magnitude of the HRI torque rises accordingly.

\begin{figure}[htb]
    \centering
    \begin{subfigure}{0.32\textwidth}
        \centering
        \includegraphics[width=\linewidth]{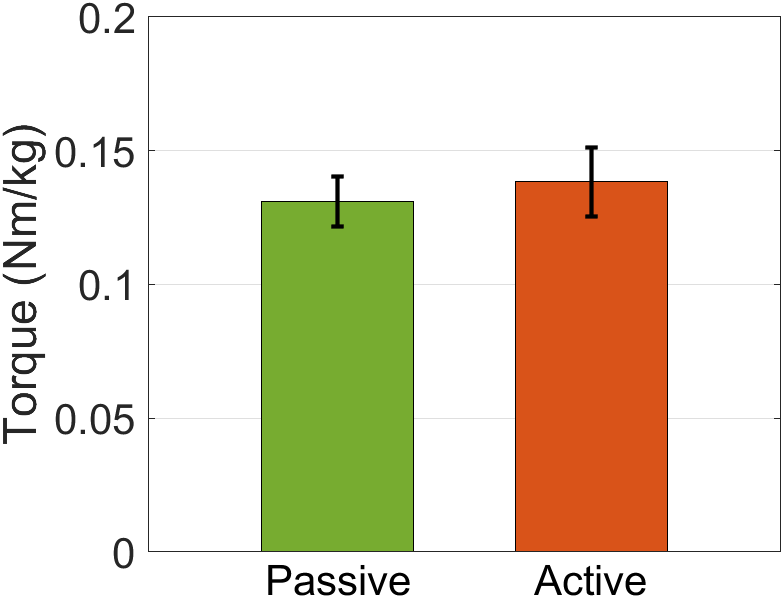}
        \caption{Maximum HRI torque.}
        \label{fig:CompareMaxHRI}
    \end{subfigure}\hfill
    \begin{subfigure}{0.32\textwidth}
        \centering
        \includegraphics[width=\linewidth]{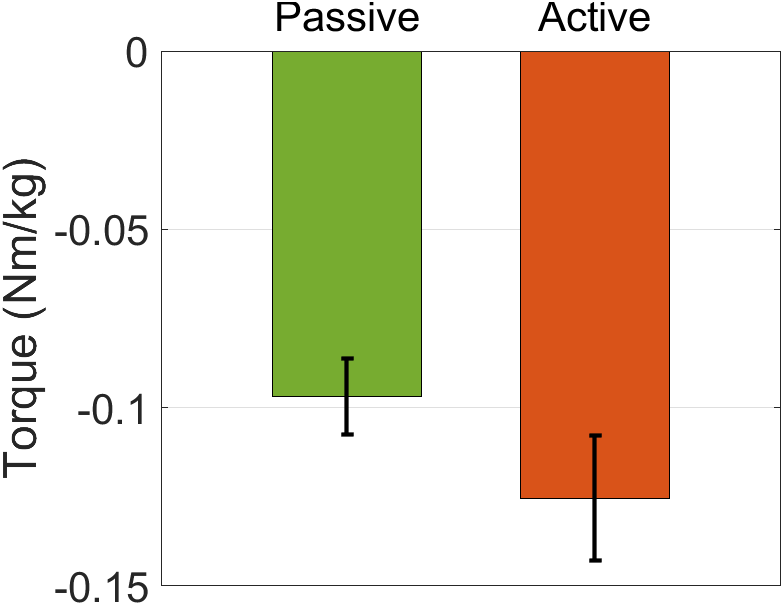}
        \caption{Minimum HRI torque.}
        \label{fig:CompareMinHRI}
    \end{subfigure}\hfill
    \begin{subfigure}{0.32\textwidth}
        \centering
        \includegraphics[width=\linewidth]{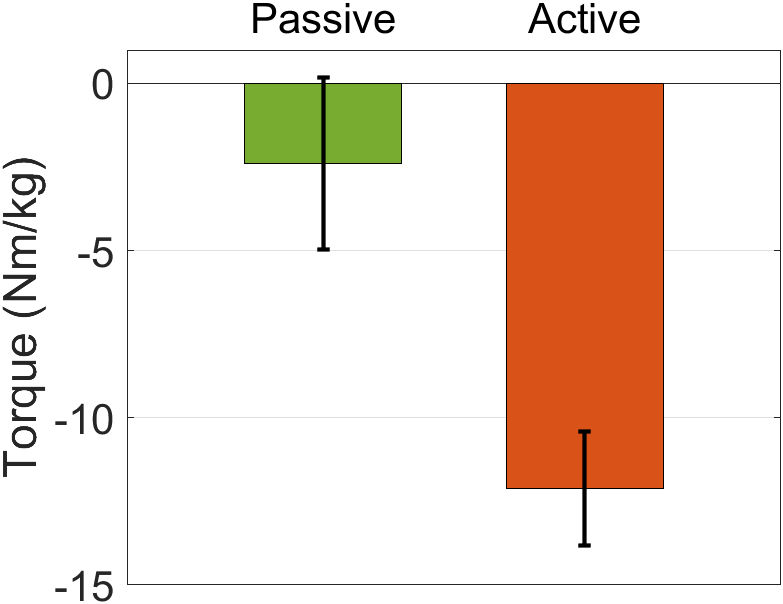}
        \caption{Cumulative HRI torque.}
        \label{fig:CompareSumHRI}
    \end{subfigure}
    
    \caption{Comparison of HRI metrics between two walking assistance conditions: passive and position.}
    \label{fig:CompareHRI}
\end{figure}

Figure \ref{fig:CompareHRI} compares the maximum, minimum and cumulative HRI torque over the gait cycles acquired in the two walking conditions.

The maximum HRI torque was detected in the early phase of the gait cycle for both control modes and presented similar values. In contrast, the minimum HRI torque was generally observed around 50\% of the cycle, showing different values depending on the control setting. In active control, the HRI was approximately 37\% more negative than in passive control, showing stronger forces in the direction of DF.
The cumulative HRI torque value was approximately one order of magnitude greater in active mode, indicating a less symmetrical and predominantly negative HRI torque profile in this control configuration.

\subsection{Load-Carrying Assistance}

The second experiment extends the analysis of the human-device complex to another dynamic task: load-carrying. 
The objectives of this study focus on several critical aspects. The first is to investigate the effect of load transport on human ankle kinematics. The second is to determine whether the muscles responsible for ankle movement exhibit increased activity during front pack transport.
This study also aims to assess whether the passive and/or active assistance modes of the exoskeleton can effectively reduce the muscular effort exerted during load-carrying tasks.
Lastly, the study seeks to compare the HRI torque under both loaded and unloaded walking conditions to understand how the additional load influences the human-robot interaction.

\subsubsection{Kinematics}
\begin{figure}[htb]
    \centering
    \begin{subfigure}[t]{0.55\textwidth}
        \centering
        \includegraphics[width=\linewidth]{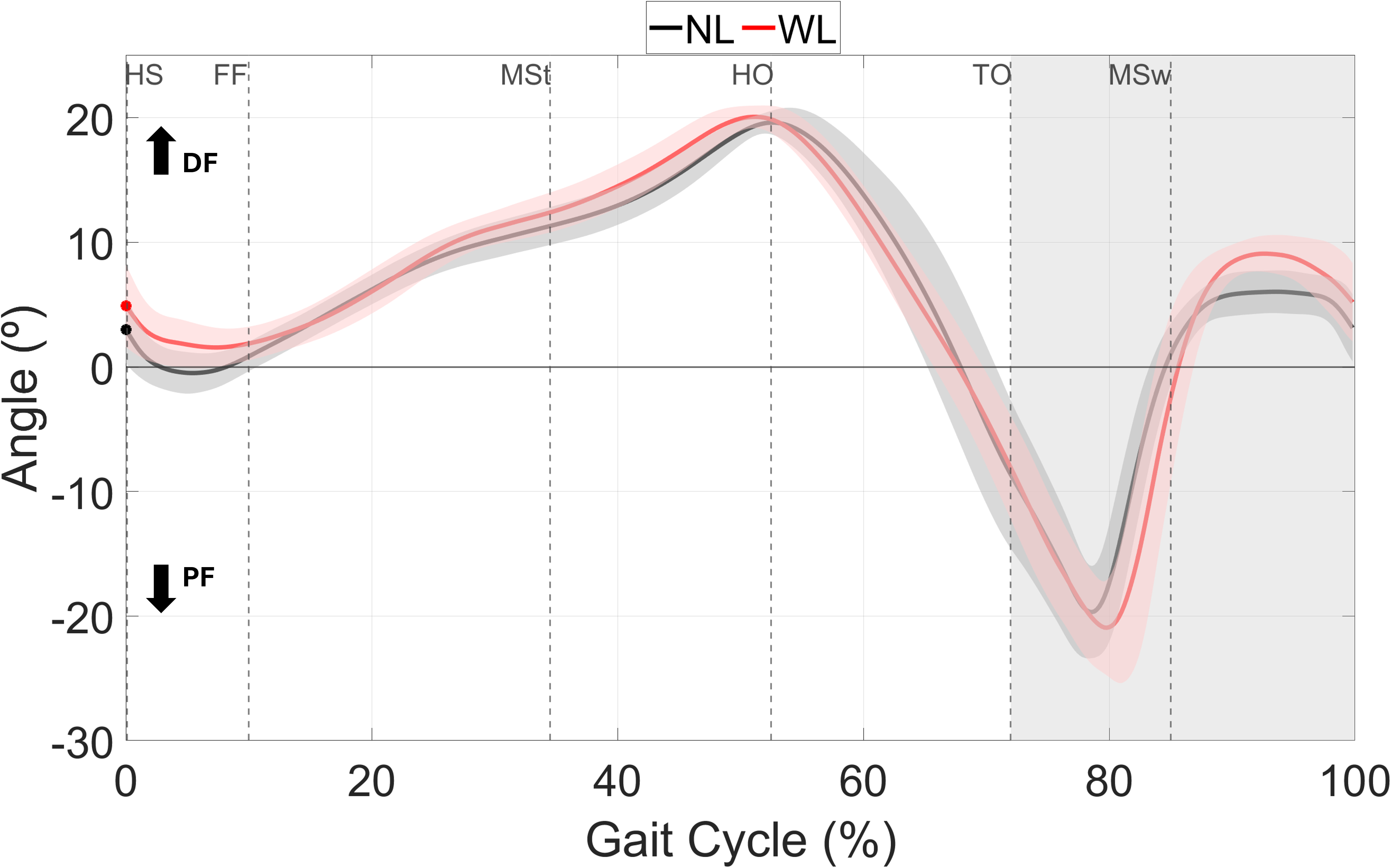}
        \caption{No assistance.}
        \label{fig:AngleNoExoCompareLoads}
    \end{subfigure}
    \vspace{0.1cm}
    \begin{subfigure}[t]{0.49\textwidth}
        \centering
        \includegraphics[width=\linewidth]{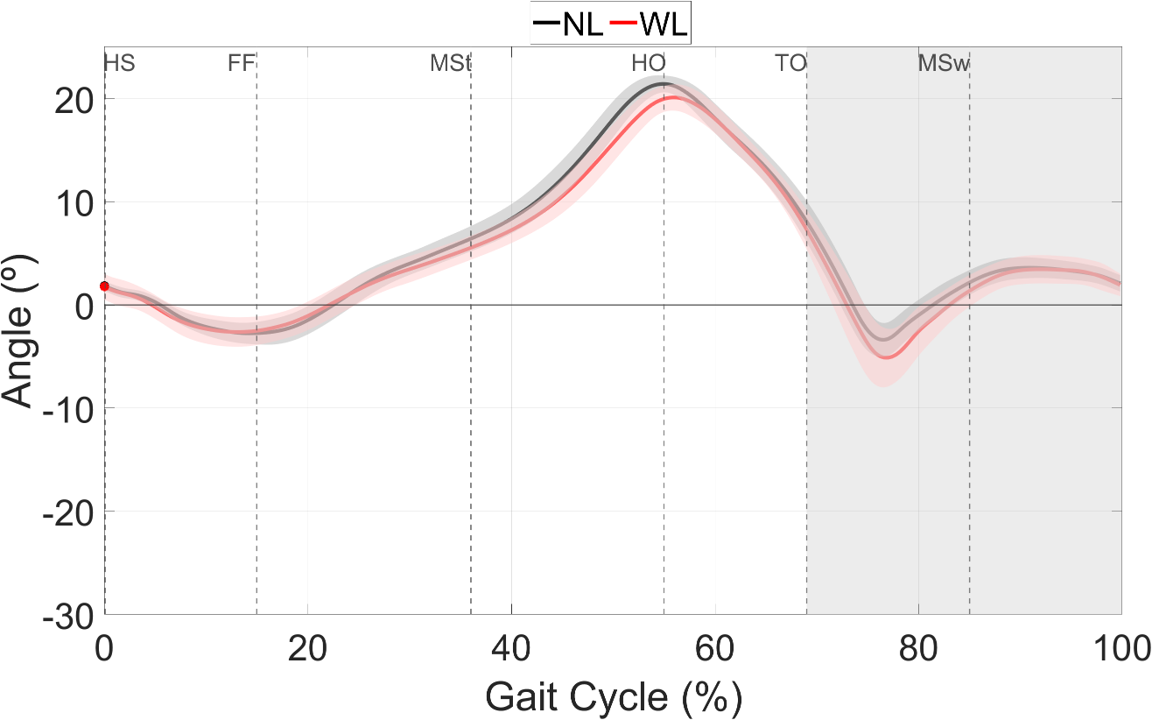}
        \caption{Passive assistance.}
        \label{fig:AngleExoPassiveCompareLoads}
    \end{subfigure}
    \hfill
    \begin{subfigure}[t]{0.49\textwidth}
        \centering
        \includegraphics[width=\linewidth]{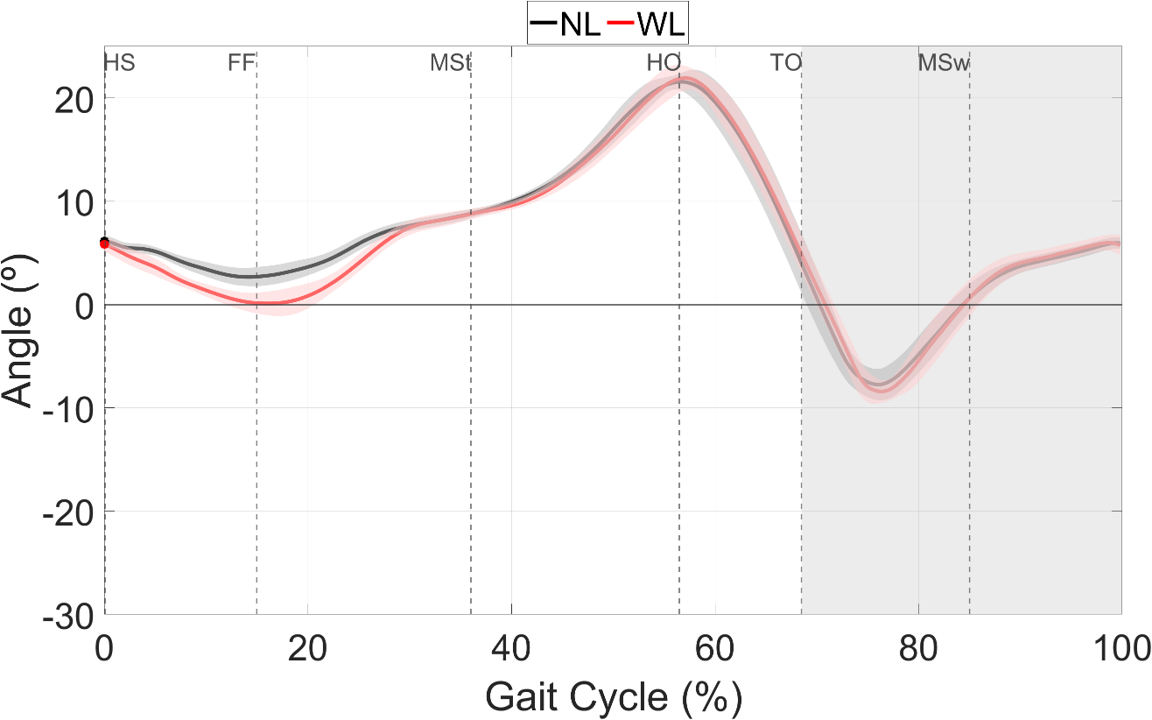}
        \caption{Active assistance.}
        \label{fig:AngleExoPositionCompareLoads}
    \end{subfigure}
    \caption{Comparison of the average angular position during one gait cycle between two loading conditions in three assistance conditions. NL - No Load; WL - With Load.}
    \label{fig:AngleCompareLoads}
\end{figure}

Figure \ref{fig:AngleCompareLoads} shows the kinematic data of the ankle joint averaged along the collected gait cycles when walking with and without loads (NL - No Loads, WL -With Load) in the three study conditions: unassisted, passive and active assistance.

Comparing the unassisted walking trials in Figure \ref{fig:AngleNoExoCompareLoads}, the results revealed no significant effect of load carrying on the maximum DF and PF angles of the ankle. The RoM during stance phase was similar in both conditions, while swing exhibited a slightly higher RoM with the load due to an increase in DF. The IC angle showed little variation due to the load. Additionally, the stance phase percentage increased when carrying the load, primarily due to an extended pre-swing phase.

During passive assistance, presented in Figure \ref{fig:AngleExoPassiveCompareLoads}, no significant differences were observed in the angular position patterns. The subject maintained consistent maximum DF and PF angles, as well as the IC angle, with no significant differences between unloaded and loaded walking trials as these metrics varied less than 2\degree.
Unlike unassisted walking, there was no increase in stance phase percentage when carrying a load.

During active position control, as shown in Figure \ref{fig:AngleExoPositionCompareLoads}, the subject's RoM and IC angle remained consistent regardless of loading conditions.
Similar to the unassisted walking observations, the stance phase percentage increased slightly when carrying loads. The figure also highlights that around FF, the subject exhibited a more pronounced PF while walking with loads, with a difference of 2.53\degree.

To complement the graphical information, Table \ref{tab:KinematicsAnkleLoads} presents kinematic metrics for comparison across the different experimental conditions.

\begin{table}[htb]
    \centering
    \footnotesize
    \caption{Descriptive analysis of human ankle joint kinematics (means and stds): NA - No Assistance, AA - Active Assistance, PA - Passive Assistance, NL - No Load, WL - With Load}
    \label{tab:KinematicsAnkleLoads}
    \begin{tabular}{ccccccc}
    \hline
     & \multicolumn{2}{c}{\textbf{NA}}   & \multicolumn{2}{c}{\textbf{PA}} & \multicolumn{2}{c}{\textbf{AA}}         \\
     & \textbf{NL} & \textbf{WL} & \textbf{NL}   & \textbf{WL}   & \textbf{NL} & \textbf{WL} \\ \hline
    \textbf{RoM (\degree)}         & 41.98 (3.24)  & 44.29 (3.16)  & 25.72 (1.73)  & 26.25 (2.49)  & 30.26 (1.32) & 31.28 (1.83) \\
    \textbf{Max DF (\degree)}      & 20.12 (0.89)  & 20.58 (0.91)  & 21.69 (0.84)  & 20.38 (1.28)  & 22.09 (0.89) & 22.33 (1.02) \\
    \textbf{Max DF (\%)}   & 52.64 (1.85)  & 50.79 (2.09)  & 55.14 (1.27)  & 55.99 (1.42)  & 56.96 (1.45) & 57.46 (1.33) \\
    \textbf{Max PF (\degree)}      & -21.86 (3.26) & -23.71 (2.98) & -4.03 (1.32)  & -5.87 (2.22)  & -8.17 (1.28) & -8.95 (0.99) \\
    \textbf{Max PF (\%)}   & 78.75 (1.38)  & 80.58 (1.52)  & 57.44 (28.90) & 72.20 (17.06) & 76.41 (1.31) & 76.68 (1.63) \\
    \textbf{Angle IC (\degree)}    & 3.00(2.73)    & 4.89 (3.35)   & 1.83 (0.51)   & 1.76 (1.80)   & 6.14 (0.56)  & 5.85 (0.75)  \\
    \textbf{Stance (\%)} & 72.10 (1.91)  & 75.24 (2.12)  & 68.54 (1.54)  & 68.98 (2.29)  & 67.06 (1.77) & 70.21 (1.75) \\ \hline
    \end{tabular}
\end{table}

\subsubsection{Muscle Activity}

\begin{figure}[htb]
    \centering
    \includegraphics[width=0.8\linewidth]{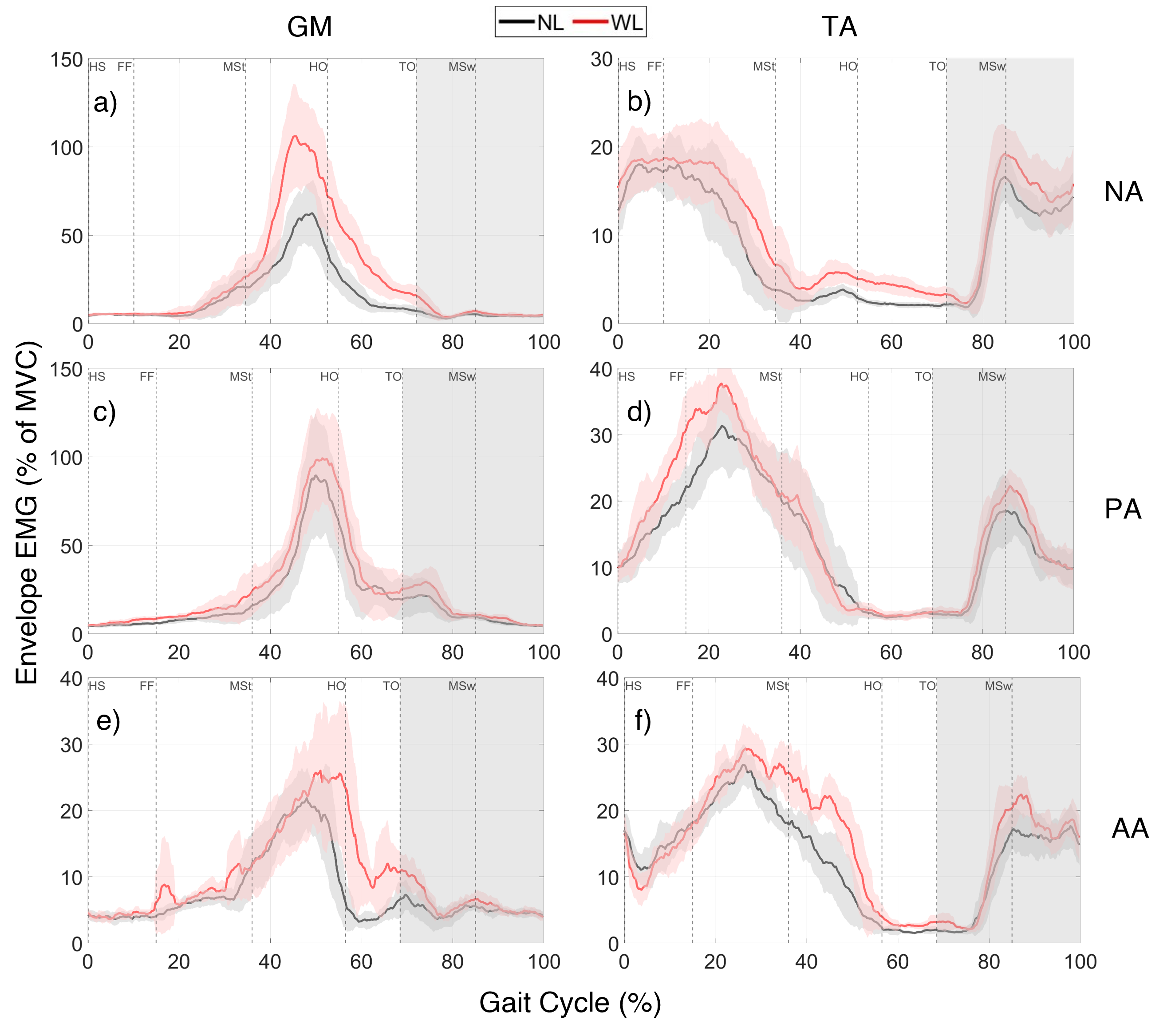}
    \caption{Comparison of average gait cycle of GM (left) and TA (right) activity in two loading conditions. NA - No Assistance (top); PA - Passive Assistance (middle); AA - Active Assistance (bottom); NL - No Load; WL - With Load.}
    \label{fig:EMGCompareLoads}
\end{figure}

Figure \ref{fig:EMGCompareLoads} shows the average muscle activity of TA and GM in the three walking conditions.

During unassisted walking, GM activation followed a similar profile regardless of load condition, but peak activation at HO increased by 70\% during loaded walking compared to unloaded walking (Figure \ref{fig:EMGCompareLoads}a).
Regarding TA (Figure \ref{fig:EMGCompareLoads}b), there seems to be an increase in its activation when walking with loads, although less evident.

Regarding muscle activity during passive assistance, the GM patterns in Figure \ref{fig:EMGCompareLoads}c show higher activation throughout the gait cycle with the load. The peak value increased by 10.8 \% for load transportation. However, the variability across the gait cycle does not reveal a clear distinction between the loaded and unloaded conditions.
For the TA, shown in Figure \ref{fig:EMGCompareLoads}d, the activation patterns are nearly identical between the two conditions, with the only noticeable difference occurring around FF and MSw, where activation increases with loads.

For active position control, Figure \ref{fig:EMGCompareLoads}e indicates that the peak activation value of the GM is, on average, higher in the loaded condition (increase of 19 \%). Although standard deviations overlap, a clearer distinction is observed in the activation profiles of GM between HO and TO.
In this assistance condition, the participant reached peak GM activation later in the gait cycle and the muscle remained with greater activation during PSw phase.
For the TA, as illustrated in Figure \ref{fig:EMGCompareLoads}f, the loaded condition resulted in noticeably greater activation between 30 and 50 \% of the gait cycle.

\begin{figure}[htb]
    \centering
    \includegraphics[width=0.8\linewidth]{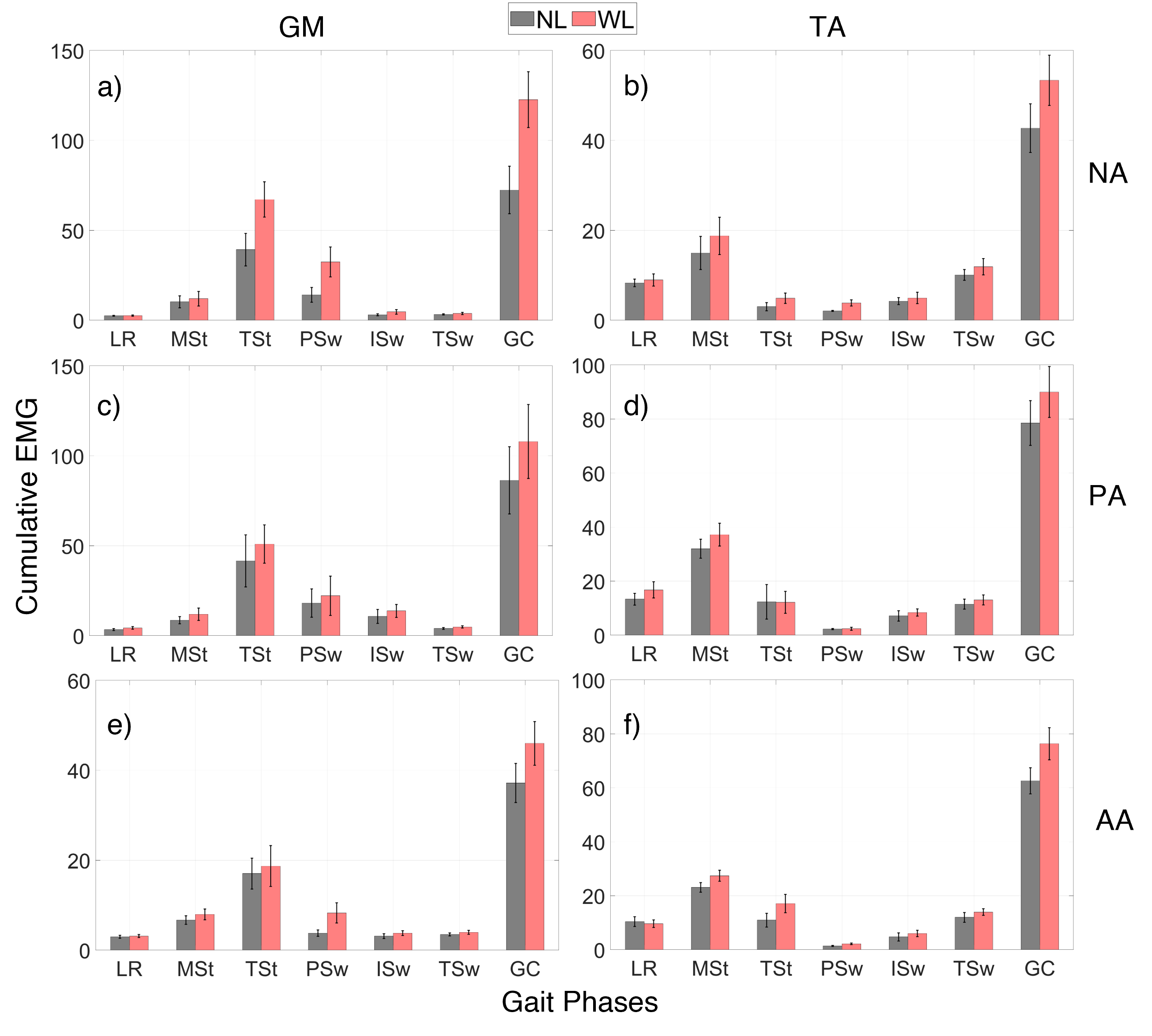}
    \caption{Comparison of cumulative GM (left) and TA (right) activity across six subphases of walking and for the complete gait cycle in two loading conditions. NA - No Assistance (top); PA - Passive Assistance (middle); AA - Active Assistance (bottom); NL - No Load (black); WL - With Load (red).}
    \label{fig:BarPlotofCumulativeEMGLoads}
\end{figure}

To allow a more thorough comparison between all the conditions in which muscle activation data was collected, Figure \ref{fig:BarPlotofCumulativeEMGLoads} shows the cumulative EMG of each muscle during the six subphases of the gait cycle and for the complete gait cycle.
Figures \ref{fig:BarPlotofCumulativeEMGLoads}a, \ref{fig:BarPlotofCumulativeEMGLoads}c and \ref{fig:BarPlotofCumulativeEMGLoads}e confirm greater GM activation when walking with loads throughout the cycle.
Significant differences were noticed in TSt and PSw for unassisted walking and for PSw in active assistance.
Furthermore, Figures \ref{fig:BarPlotofCumulativeEMGLoads}b, \ref{fig:BarPlotofCumulativeEMGLoads}d and \ref{fig:BarPlotofCumulativeEMGLoads}f reveal overall greater TA activation during load-carrying conditions.
During unassisted walking, significant increases in TA activation were observed in TSt and PSw phases of the gait cycle.
In active assistance, higher activation was evident not only during these same two phases but also during MSt.
In contrast, during passive assistance, the results were inconclusive, as no clear differences in TA activation were observed between conditions.

\subsubsection{Human-Robot Interaction}

\begin{figure}[htb]
\centering
    \begin{subfigure}{0.49\textwidth}
        \includegraphics[width=\linewidth]{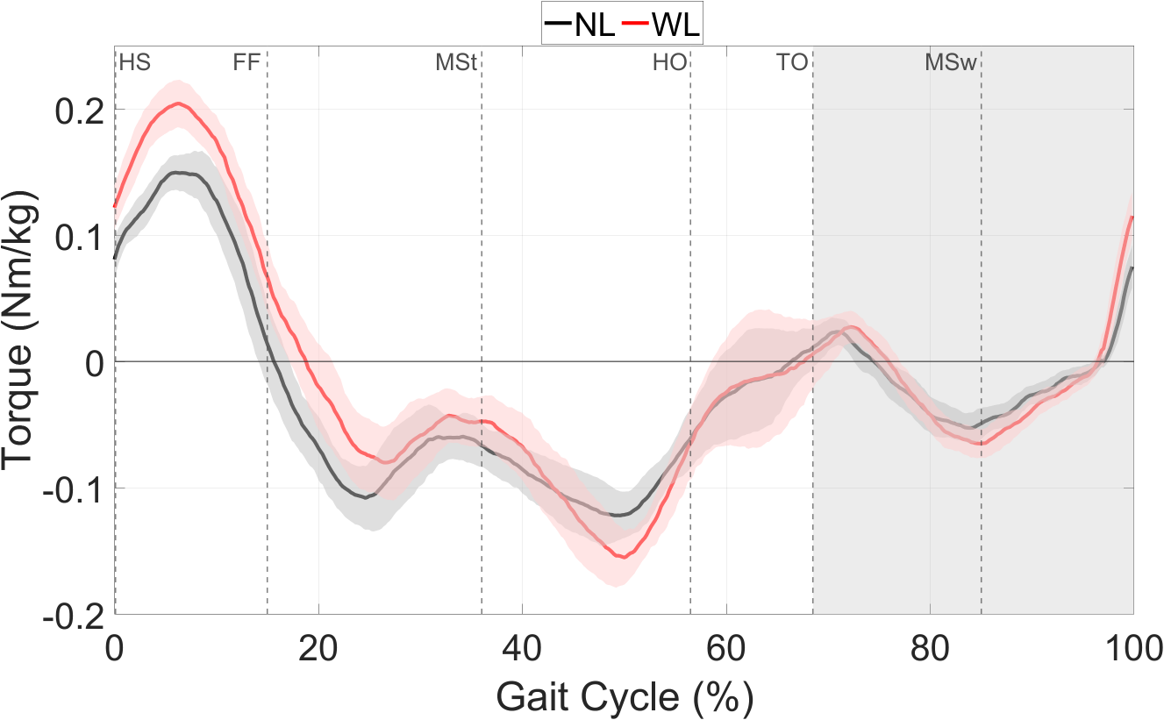}
        \caption{Active assistance.}
        \label{fig:HRIExoPositionComparisonLoads}
    \end{subfigure}
    \hfill
    \begin{subfigure}{0.49\textwidth}
        \includegraphics[width=\linewidth]{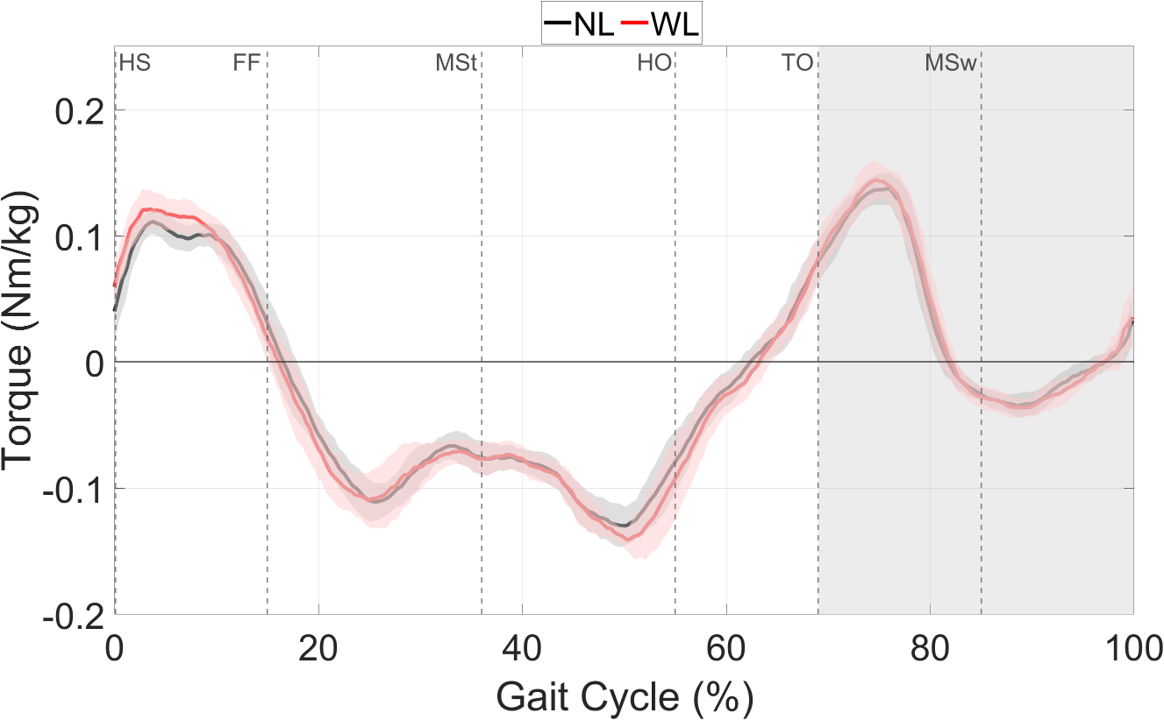}
        \caption{Passive assistance.}
        \label{fig:HRIExoPassiveComparisonLoads}
    \end{subfigure}
    \caption{Average gait cycle of the measured HRI torque in two loading conditions during passive and active assistance. NL - No Load; WL - With Load.}
    \label{fig:HRIComparisonLoads}
\end{figure}

Figure \ref{fig:HRIComparisonLoads} shows the average gait cycle of the HRI torque in two different loading conditions during exoskeleton passive and active assistance.

The interactions in passive assistance showed no difference between walking with and without loads (Figure \ref{fig:HRIExoPassiveComparisonLoads}).
However, in active assistance, the data reveals a distinction from HS to FF (LR phase) and also around 50\% of the gait cycle, preceding HO. These segments represent the moments of increased interaction between the user and the robotic device, either in the direction of PF (positive HRI) or DF (negative HRI).
During LR, both conditions recorded the maximum HRI value, being 0.168 $\pm$ 0.015 Nm/kg without loads and 0.211 $\pm$ 0.018 Nm/kg with loads. Between FF and HO, the minimum HRI values were recorded, being -0.135 $\pm$ 0.020 Nm/kg without loads and -0.169 $\pm$ 0.017 Nm/kg with loads. 
In both segments, the HRI torque exhibits stronger interactions (higher in module) under the loaded walking condition.

\begin{figure}[htb]
    \centering
    \begin{subfigure}{0.49\textwidth}
        \includegraphics[width=\linewidth]{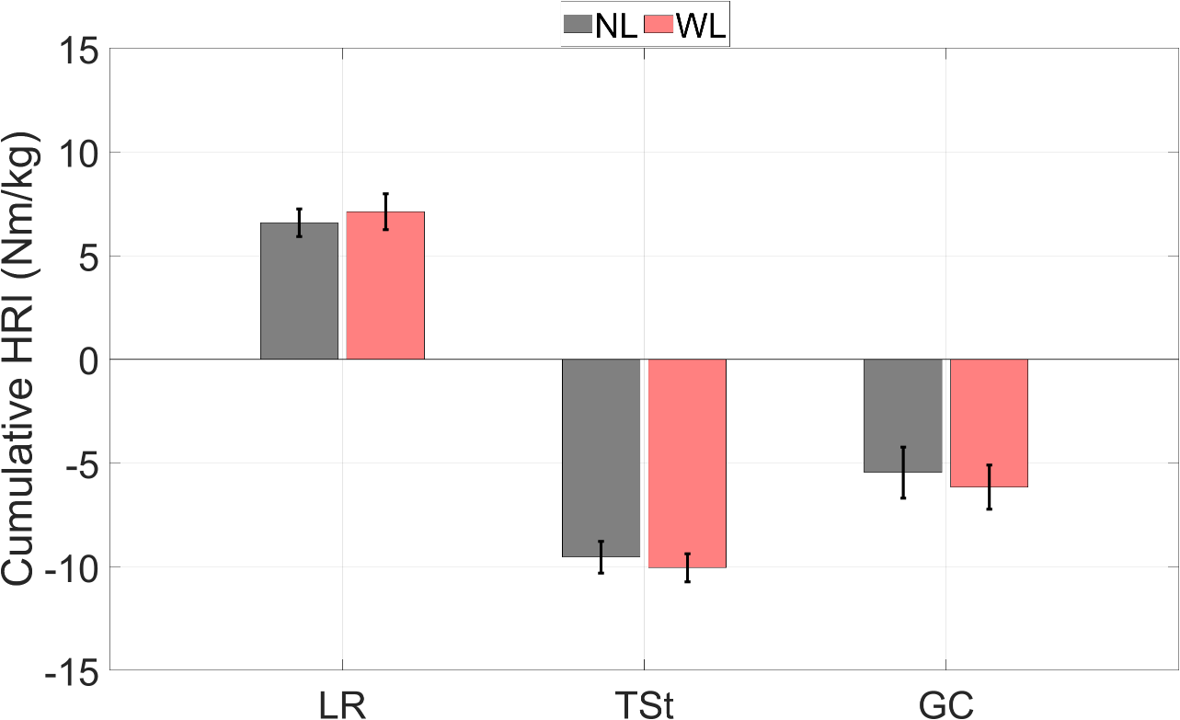}
        \caption{Passive Assistance.}
        \label{fig:CumulativeHRIPassiveLoads}
    \end{subfigure}
    \hfill
    \begin{subfigure}{0.49\textwidth}
        \includegraphics[width=\linewidth]{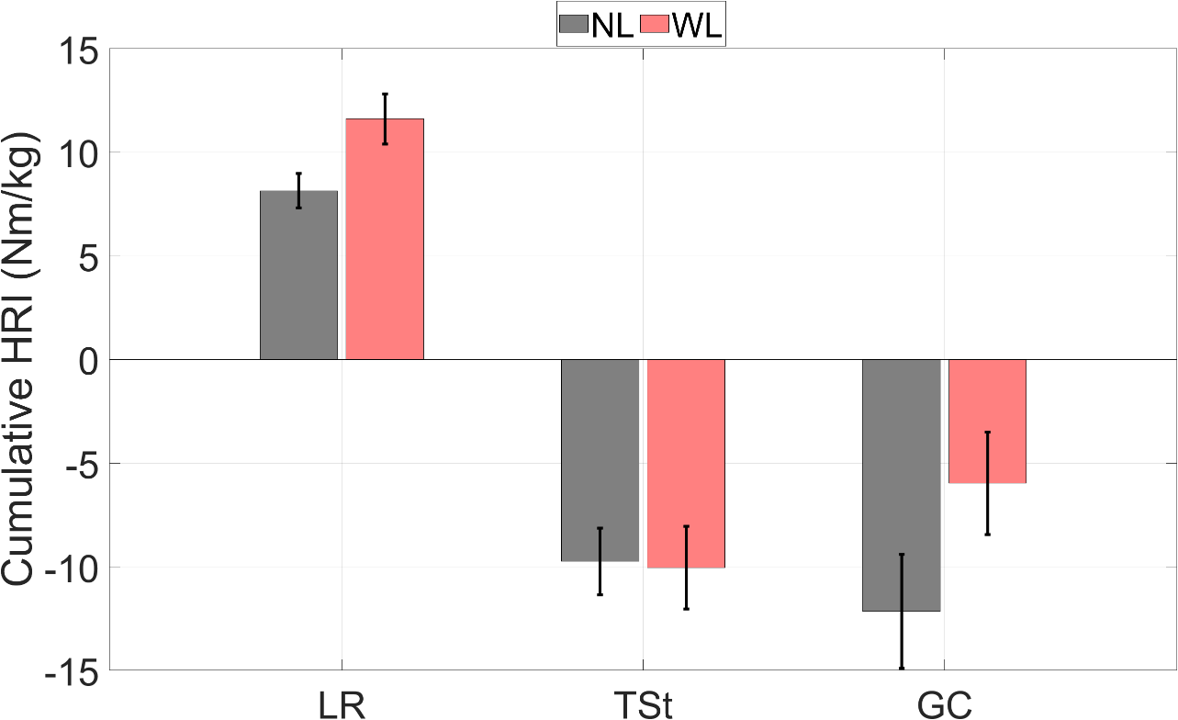}
        \caption{Active assistance.}
        \label{fig:CumulativeHRIPositionLoads}
    \end{subfigure}
    \caption{Comparison of cumulative HRI torque in two gait phases (LR and TSt) and throughout the gait cycle (GC).}
    \label{fig:CumulativeHRIMetricsLoads}
\end{figure}

Figure \ref{fig:CumulativeHRIMetricsLoads} presents the cumulative HRI torques during two gait phases and across the entire gait cycle to compare the loading conditions. LR occurs between HS and FF and TSt occurs between MSt and HO.
In passive control, the cumulative results similarly reveal no significant differences between the loading conditions.

In active assistance mode, the LR phase presents significantly higher interactions while carrying a load. On the contrary, for the TSt phase, the cumulative values do not show a notable increase in interactions due to the load. While the peak HRI torque may be higher, the overall interactions remain identical.
Regarding the cumulative HRI over the gait cycle, the less negative value in the loaded condition compared to the unloaded condition suggests that the interactions were more symmetrical, although still with a tendency to move the device towards DF.

\section{Discussion}

\subsection{Basic Ankle Movements}

This protocol allowed to understand that the HRI torques applied in the direction of DF record negative interaction values. Symmetrically, HRI torques that generate PF movement result in positive interaction values.
Data revealed that the magnitude of the HRI torque depends on the intensity of the force that the user applies through the straps.

In addition, it was found that the gravitational component, i.e. the weight of the person acting on the device, can produce different interaction torques (positive or negative) depending on the movement being executed (PF or DF). In some cases, the weight aligns and helps the movement of the device, while in others, it acts as an opposing force.

These findings align with indications that minimal interaction torques suggest synchronized movement between the human and the robot \citep{pons2008wearable}, as neither the device pushes against the user's foot nor the user exerts force against the device.
When the user applies voluntary forces, they are detected as increased HRI torque magnitudes, with the torque's sign offering insight into the direction of the user's intended movement.

Additionally, the HRI torque provides insights into the user's comfort. An increase in the magnitude of interactions reflects higher forces exerted between the user and the exoskeleton, which can result in increased contact pressure and discomfort, thereby diminishing the overall experience of using the device. While comfort is inherently challenging to quantify, it is well-established that minimizing interaction forces enhances the user's comfort.

\subsection{Ankle Assistance}

\subsubsection{Gait Kinematics}
In general, the mean angular position of the ankle joint during walking matches other literature findings \citep{grimmer2020human, reznick2021lower, wang2023effect}. The stance phase was approximately 71\% of the gait cycle, exceeding the typical 60-65\% due to the slower walking speed of 1.3 km/h \citep{hebenstreit2015effect}.

Comparing the RoM of the participant's foot with and without assistance, it was found that there was a slight limitation of RoM with the exoskeleton, particularly due to differences in PF.
In passive mode, a greater kinematic variability between subjects was verified that can be attributed to the way the device is attached to the user's leg.
The fixed strap position may lead to less effective coupling for some participants due to biomechanical differences, making it harder to move the device into desired positions.
In active position control, foot movement was constrained considering that the pattern imposed by the device had a limited RoM ([-11 \degree, +12 \degree]).
Even so, it is considered that users were able to replicate a normal gait pattern and walk efficiently while wearing the device in any control setup.

In active control, the user's foot contacted the ground earlier due to the device's added weight, possibly causing swing instability, and the lack of adaptation to the user's stride variability, hampering synchronization. In passive mode, delays were expected as the exoskeleton mirrors user interactions, with its inertia causing a retarded reaction.

Kinematic data also demonstrated that the device's RoM was lower than the one of the user's during walking.
In passive control, it indicates that the exoskeleton cannot be completely complaint to the user's interactions.
In position control, it demonstrates that the foot retains freedom of movement in parallel to the pattern imposed by the device, meaning that the user's movement is not only device-driven.

\subsubsection{Muscle Activity}

In general, the collected EMG of the GM and TA is consistent with other literature findings \citep{kwak2023fascicle}.
GM activates during terminal stance to propel the body forward, while TA stabilizes the foot at initial contact and lifts it in swing to prevent dragging.

Since the ankle exoskeleton is designed as a gait assistance device, it would be expected to reduce the users' muscle activation.
For GM, there was a significant reduction when walking with the robotic device, more evident in active control.
This would be the result of the aid of the exoskeleton as well as the reduction of the maximum PF angle indicated by the analysis of the kinematic data.

The effects of the device on TA muscle engagement were less pronounced. However, analysis of the complete gait cycle reveals that muscle activation was reduced more in passive mode than in active position mode, differently from GM.
This could be because, in active mode, the exoskeleton's movements may introduce some instability, prompting greater activation of the TA to maintain balance, whereas movements in passive control are more predictable as they are driven by the user. Overall, the less significant reduction on the TA might also be attributed to the unilateral design of the exoskeleton, which requires additional effort to maintain balance during walking.

\subsubsection{Human-Robot Interaction}

The HRI patterns observed during passive and active assistance demonstrate the forces imposed by the user on the ankle exoskeleton.
In passive control, it can be seen that the participants pushed the device in the same direction as the movement of their foot, thus presenting positive interactions when doing PF and negative interactions when doing DF.
In active control, the HRI throughout the gait cycle is similar to passive control, with a distinction during push-off.

In passive control, during the push-off phase, the user must actively exert force to drive the exoskeleton into PF, resulting in a positive HRI torque peak near TO. In contrast, during active control, the push-off phase exhibits a negative HRI torque that reaches zero at TO. This indicates that, in this mode, the exoskeleton pushes the user's foot to propel the body forward. At TO, the user eventually compensates for its own weight to move the leg during the swing phase.

In both control modes, a positive interaction peak is observed during LR phase. In passive mode, this peak occurs because the user’s foot pushes the device into PF. In active mode, the exoskeleton should perform PF after HS. While it is expected that the HRI torque would still be positive due to the participant’s weight acting on the insole, the magnitude should be smaller. However, the observed HRI torque values highlight a lack of synchronization between the movements of the user and the exoskeleton. This misalignment reflects lack of adaptability.

Additionally, in active control, a high negative HRI torque is observed before HO. This indicates insufficient assistance in performing DF, as the user is compelled to push the lower limb in that direction. These findings suggest that by detecting elevated interaction torques and cross-referencing them with the movements of both the user and the exoskeleton, it becomes possible to identify moments in which the user’s needs are not being adequately met, revealing opportunities for improvement in control design.

The results align with findings in the literature, which suggest that reduced interaction torques are indicative of improved synchronization between the human and the robotic device. 

\subsection{Load Carrying}

\subsubsection{Kinematics}

The findings regarding the RoM of the ankle joint during unassisted walking are consistent with other studies examining load-carrying tasks, as they also indicate little to no effect due to load \citep{wang2023effect, dames2016effects}. 
This analysis confirmed that even under passive and active assistance with a robotic ankle device, the RoM remained unchanged. This indicates that load transport has no significant effect on the kinematics of the human ankle.

Additionally, a percentage increase in the stance phase was observed when comparing all conditions with and without load, more evidenced during unassisted walking. This observation aligns with existing literature \citep{dames2016effects, harman2000effects}, which notes that carrying a load alters the body's balance, and to maintain a more stable movement, the duration of the double support phases increases. The further away the load is positioned from the body's Centre of Mass (CoM), the greater the instability introduced, leading to a more pronounced effect on gait phases. In this case, since the load was placed in front of the participant and close to the body, the change in the stance phase percentage was minimal across all conditions.
In active position control, differences were minimal because the robotic device enforces a pre-defined pattern. So, if the user intended to increase the double support time, the device would not accommodate such adjustments, instead maintaining the fixed positions dictated by its programmed trajectory.

\subsubsection{Muscle Activity}

This study provides evidence of increased muscle activation for both GM and TA during loaded walking. These findings are consistent with other literature \citep{wang2023effect, dames2016effects, silder2013men, genitrini2022impact}. 
Carrying a load demands the muscles to work harder to maintain balance, generate additional force for propulsion (GM), and ensure smooth contact with the ground (TA).

With exoskeleton assistance, the load still caused an increase in muscle activity, although not as evident as without the device. The load's impact was lower with passive assistance than with active assistance. However, since absolute activation values were significantly higher in passive mode, it cannot be concluded that this mode is more suitable for load-carrying assistance.

These findings suggest that the ankle exoskeleton, although capable of reducing muscle activity compared to unassisted walking, does not fully compensate for the effort required by the subject to carry a 10 kg load.
Nonetheless, the differences in GM muscle activity in the TSw phase indicate that the device can provide partial compensation.  
For full efficiency, the control would need to be customised to the specific task of load carriage.

\subsubsection{Human-Robot Interaction}
The results of this study contribute to the literature by revealing that the interaction torque between the ankle exoskeleton and the user can present changes due to the additional loads placed on the human body while walking.

Since HRI torque is influenced by the movement of the robotic device and the user's lower limb, passive and active control disclosed different outcomes.
Under passive control, the HRI torque was not significantly affected by the load. 
This control mode adjusts the device's movement based on the user's interactions. The exoskeleton responds to the user's force, moving when sufficient input is applied to overcome its inertia and maintain controlled torque levels, therefore preventing the HRI torque from exceeding certain values. 
While the interaction minimization is a positive feature of passive control, it limits the assistance provided to the user, offering only mechanical support without compensating for the additional effort required to transport the load.

Differently, position control applies a pre-defined movement pattern that does not adapt to the user’s needs over time. As a result, carrying a 10 kg load leads to increased HRI torques at specific phases of the gait cycle, particularly when the device cannot adequately assist the user’s joint during critical moments. On that account, during LR the HRI torque reaches higher magnitudes. 
Around 50 \% of the gait cycle, an increase in negative HRI torque is observed, preceding HO. This is likely an anticipatory response by the user, who appears to push the ankle exoskeleton into DF to prepare for forward propulsion during the pre-swing phase. This phase becomes particularly challenging under load-carrying conditions, as the body requires additional strength for push-off.

This analysis demonstrated that HRI torque is affected when the user is carrying loads. If the control system does not adequately respond to the interactions applied by the user, more intense interaction torques are generated. The increase of HRI torque whether positive or negative can serve as an indicator of the user's need for greater adaptability and enhanced assistance to perform the motor task.

\section{Conclusions}

To conclude, this research revealed insights into the ankle exoskeleton's potential for aiding locomotion and load transport. 
Walking with the device resulted in a slight reduction in ankle RoM. However, a normal movement pattern was preserved. Muscle activity of GM and TA decreased during gait with the device. Regarding the effects of front packs, ankle joint kinematics showed no significant changes, but increased engagement of the shank muscles was observed during this activity.
The study revealed that users can exhibit different behaviours when walking with the same device. This underlines the uniqueness of each individual's gait dynamics, reaffirming that robotic control should assist-as-need to effectively accommodate human natural variability.
One of the most important contributions of this analysis lies in the study of the HRI. The findings highlight the importance of HRI as feedback for enhancing control adaptability by assessing the user's needs in real time, aiming to minimize interactions while delivering effective assistance.
This biomechanical analysis serves as a foundation for developing control strategies that provide adaptive and personalized assistance for walking and load-carrying tasks.
Establishing this type of analysis as a standard enables meaningful comparisons with other devices and strategies, advancing the field of assistive technology by prioritizing user-centric performance.


\newpage
\appendix
\section{Placement of Sensors}
\label{app1}
\begin{figure}[h]
    \centering
    \includegraphics[width=0.5\linewidth]{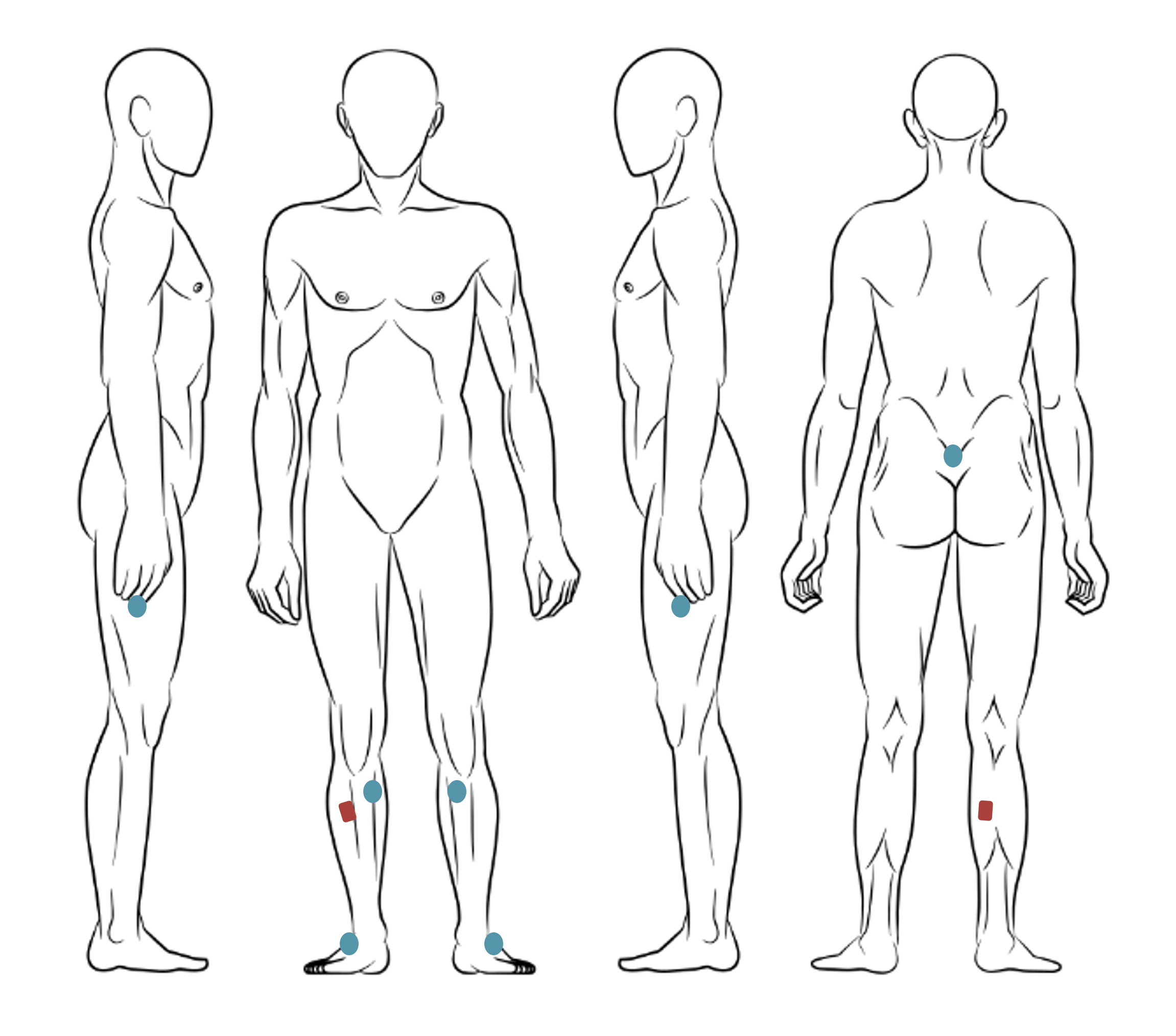}
    \caption{Sensors Placement for data collection: Xsens IMUs (blue dots) and Delsys EMG electrodes (red squares).}
    \label{fig:sensorsplacement}
\end{figure}

Figure \ref{fig:sensorsplacement} illustrates the placement of the Xsens and Delsys sensors on the participants' bodies. Blue dots indicate the locations of the IMUs, while red squares represent the EMG sensors.

\section{Diagrams of Protocols}
\label{app2}

\begin{figure}[h]
    \centering
    \includegraphics[width=0.8\linewidth]{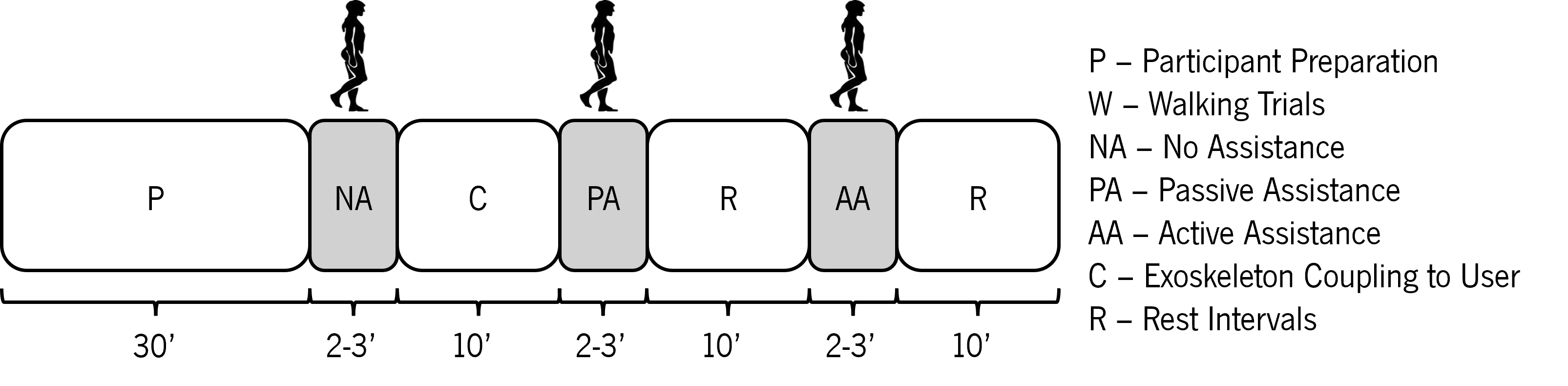}
    \caption{Schematic diagram of ankle assistance protocol.}
    \label{fig:Protocol1}
\end{figure}

\begin{figure}[h]
    \centering
    \includegraphics[width=0.8\linewidth]{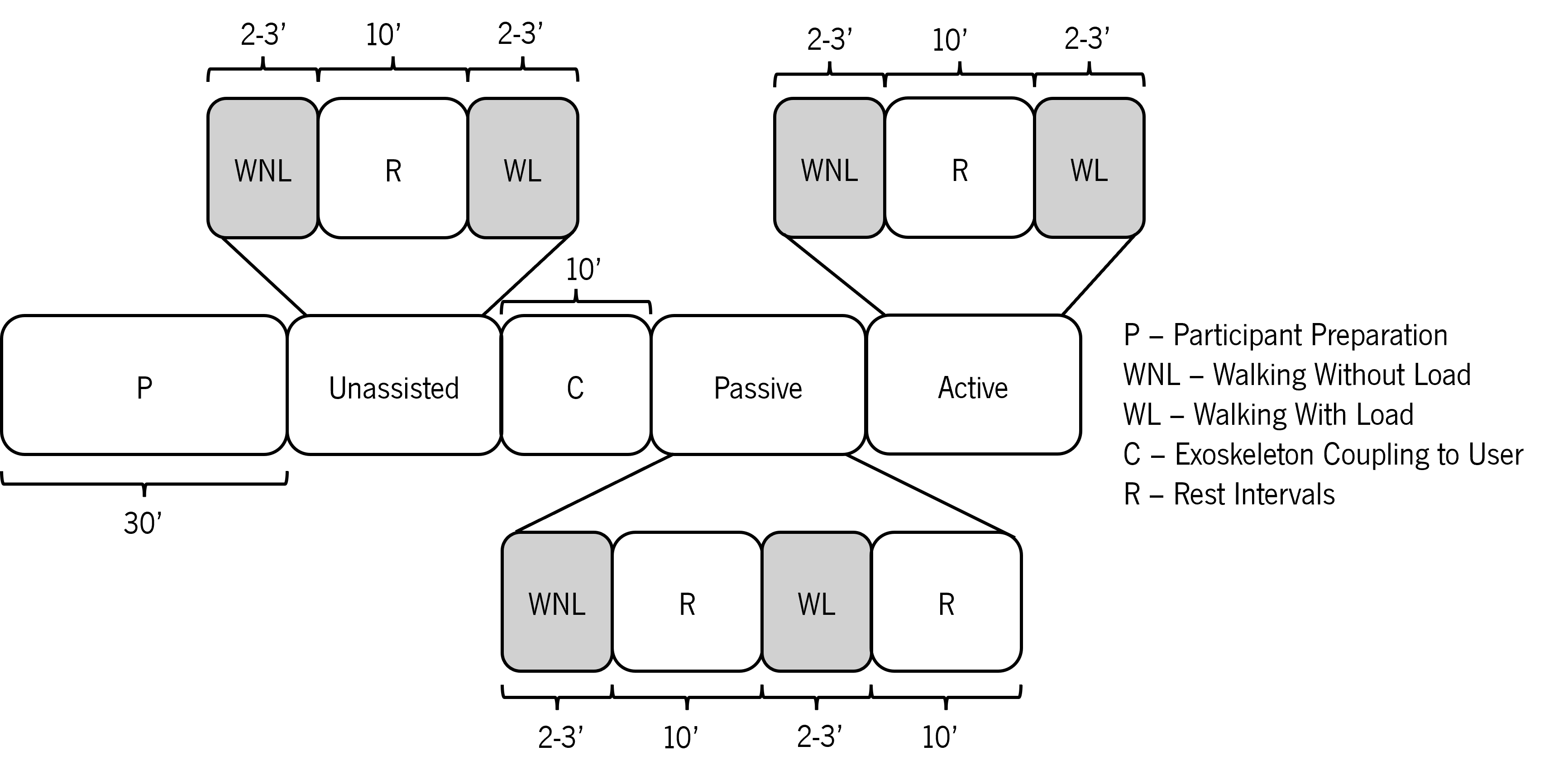}
    \caption{Schematic diagram of load-carrying protocol.}
    \label{fig:Protocol2}
\end{figure}

Figure \ref{fig:Protocol1} illustrates the timeline of the ankle walking assistance protocol, while Figure \ref{fig:Protocol2} presents the task diagram for the load-carrying protocol.

\end{document}